\documentclass{article}

\PassOptionsToPackage{numbers, compress}{natbib}

\usepackage[final]{neurips_2025}

\usepackage[utf8]{inputenc} 
\usepackage[T1]{fontenc}    
\usepackage{hyperref}       
\usepackage{url}            
\usepackage{booktabs}       
\usepackage{amsfonts}       
\usepackage{nicefrac}       
\usepackage{microtype}      
\usepackage[table]{xcolor}         

\usepackage{subcaption}
\usepackage{makecell}
\usepackage{physics}
\usepackage[capitalize,noabbrev]{cleveref}
\usepackage{xspace}
\usepackage{amsmath}
\usepackage{amssymb}
\usepackage{mathtools}
\usepackage{amsthm}
\usepackage{thmtools, thm-restate}
\usepackage{cancel}
\usepackage{algorithm}
\usepackage{algorithmic}
\usepackage{wrapfig}
\usepackage{multirow}

\usepackage{mdframed}
\usepackage{listings}
\usepackage{spverbatim}
\theoremstyle{plain}

\theoremstyle{definition}

\theoremstyle{remark}

\title{Incentivizing LLMs to Self-Verify Their Answers}

\author{%
  Fuxiang Zhang\textnormal{\textsuperscript{1,2}}\\
  \And
  Jiacheng Xu\textnormal{\textsuperscript{1,2}}\\
  \And
  Chaojie Wang\textnormal{\textsuperscript{2}}\\
  \And
  Ce Cui\textnormal{\textsuperscript{2}}\\
  \And
  Yang Liu\textnormal{\textsuperscript{2}}\\
  \And
  Bo An\textnormal{\textsuperscript{1,2}}\thanks{Corresponding author. Emails: \texttt{fuxiang001@e.ntu.edu.sg}, \texttt{jiacheng005@e.ntu.edu.sg}, \texttt{\{chaojie.wang,ce.cui,kee.liu\}@kunlun-inc.com}, \texttt{boan@ntu.edu.sg}}\\
  \AND
  \textnormal{\textsuperscript{1} Nanyang Technological University, Singapore} \quad \textnormal{\textsuperscript{2} Skywork AI}
}

\begin{document}

\maketitle

\begin{abstract}
Large Language Models (LLMs) have demonstrated remarkable progress in complex reasoning tasks through both post-training and test-time scaling laws. While prevalent test-time scaling approaches are often realized by using external reward models to guide the model generation process, we find that only marginal gains can be acquired when scaling a model post-trained on specific reasoning tasks. We identify that the limited improvement stems from distribution discrepancies between the specific post-trained generator and the general reward model. To address this, we propose a framework that incentivizes LLMs to self-verify their own answers. By unifying answer generation and verification within a single reinforcement learning (RL) process, we train models that can effectively assess the correctness of their own solutions. The trained model can further scale its performance at inference time by verifying its generations, without the need for external verifiers. We train our self-verification models based on Qwen2.5-Math-7B and DeepSeek-R1-Distill-Qwen-1.5B, demonstrating their capabilities across varying reasoning context lengths. Experiments on multiple mathematical reasoning benchmarks show that our models can not only improve post-training performance but also enable effective test-time scaling. Our code is available at \url{https://github.com/mansicer/self-verification}.
\end{abstract}

\begin{figure}[ht]
  \centering
  \includegraphics[width=0.95\textwidth]{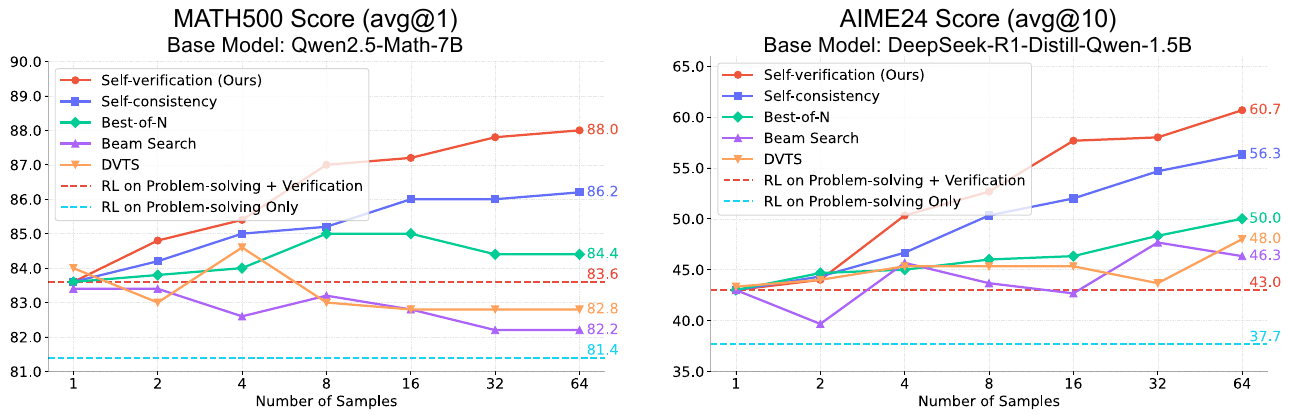}
  \vspace{-5pt}
  \caption{Average performance of post-trained models (dotted lines) and test-time scaling methods (solid lines) on the MATH500 and AIME24 benchmarks. Our self-verification framework not only enhances post-training performance with RL on both problem-solving and verification, but also enables effective test-time scaling with increased generation numbers by verifying its own solutions.}
  \label{fig:scaling-figure}
  \vspace{-9pt}
\end{figure}

\section{Introduction}

\begin{figure}[t]
  \centering
  \includegraphics[width=\textwidth]{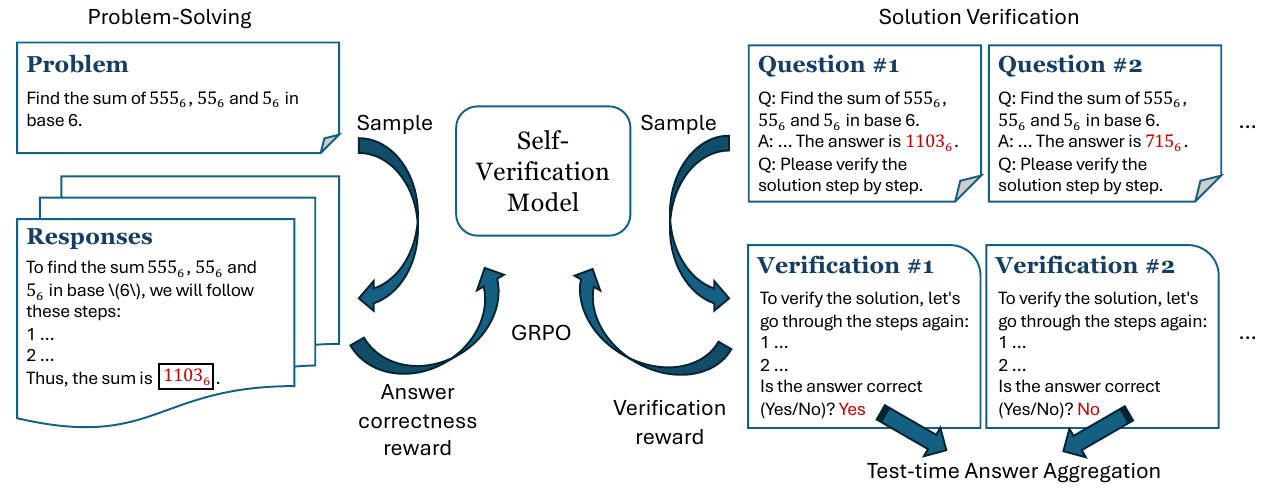}
  \vspace{-18pt}
  \caption{The framework of our self-verification framework. The model is trained to solve mathematical reasoning problems and verify generated solutions simultaneously.}
  \label{fig:intro-framework}
\end{figure}

Large Language Models (LLMs) have demonstrated remarkable capabilities in a wide range of natural language tasks \citep{instructgpt,gpt-4o,claude-37-sonnet}, particularly excelling in complex reasoning challenges such as mathematics and logic puzzles \citep{math-dataset, logic-rl}. For these verifiable tasks with ground-truth answers, researchers have proposed scaling laws in both post-training and test-time reasoning that substantially enhance LLM performance \citep{openai2024reasoning}. On the one hand, reinforcement learning (RL) has been widely adopted to align LLMs by rewarding predictions that match gold answers \citep{deepseek-r1}. On the other hand, researchers leverage additional computational resources at inference time to improve accuracy through ensemble approaches \citep{self-consistency, more-agents} and dedicated reward models (RMs) as verifiers \citep{prm-openai, process-outcome, process-evaluator}.

Although both post-training and test-time scaling laws have significantly advanced the performance of LLMs, we observe that their combination yields limited synergistic benefits. Post-trained models are often specifically tuned on training data, resulting in a narrow generation distribution, which is often unrecognized by external verifiers trained on regular data. This distribution shift may lead to incorrect solution verifications and thus hinders test-time scaling performance \citep{imperfect-verifiers}. As illustrated in \cref{fig:scaling-figure}, we find that commonly used test-time scaling methods such as best-of-N and beam search with external RMs offer minimal improvements compared to the simple self-consistency method through majority voting. It remains challenging to synergize post-training and test-time scaling within an effective framework.

In this work, we propose to leverage \textit{self-verification} to bridge the gap between post-training and test-time scaling. As illustrated in \cref{fig:intro-framework}, the core idea of self-verification is straightforward: we train the model not only to generate answers but also to verify the correctness of its solutions. Inspired by previous generative verifier works \citep{generative-verifier, generative-rm, inference-time-rm}, we treat the pretrained LLM as a verifier, and then adopt RL algorithms to incentivize the LLM for self-verification, using reward signals derived from ground-truth answer correctness. We further design an online policy-aligned buffer and dynamic verification rewards to stabilize the joint training of answer generation and verification. The online buffer ensures that the input distribution of the verifier remains aligned with the latest model outputs, while the dynamic reward function utilizes multiple rollouts from the GRPO algorithm \citep{deepseek-math, deepseek-r1} to automatically adjust reward signals, optimizing performance on challenging verification tasks. At inference time, we leverage the post-trained model for both answer generation and verification and adopt a weighted answer aggregation based on verification scores. Thanks to the unified model, our test-time scaling approach can be easily deployed within existing LLM inference engines, without relying on external RMs. Our experiments on mathematical reasoning benchmarks demonstrate that incorporating self-verification into post-training not only improves model performance but also enables effective test-time scaling with increasing generation budgets. 

\section{Preliminaries}

\subsection{GRPO for Math Reasoning Tasks}
\label{sec:grpo}

Reinforcement Learning (RL) has proven to be an effective approach for post-training a pretrained LLM with feedback from humans \citep{instructgpt}, AI proxies \citep{rlaif, rlhf-rlaif}, or ground-truth verification \citep{wizardmath, deepseek-math}. To model the LLM generation process as a Markov Decision Process (MDP), we denote the language model as a policy $\pi_\theta(y \mid x)$, where $\theta$ is the model parameters, $x$ is the input query and $y=\{y_1, y_2, \ldots, y_{T}\}$ is the generated output sequence. The policy $\pi_\theta$ can be decomposed into a sequence of conditional probability distributions over tokens $\pi_\theta(y \mid x) = \prod_{t=1}^{T} P(y_t \mid y_{<t}, x; \theta)$, given the previous tokens.

To train policy $\pi_\theta$ with RL, we also need to define a reward function to provide training signals. For our math reasoning tasks, we follow the previous DeepSeek-R1 \citep{deepseek-r1} work to adopt the correctness of the generated answer as the reward, evaluated by a rule-based verifier. Suppose the ground-truth answer to problem $x$ is $a^*$. The correctness reward $r_c$ can be defined as $r_c = \mathbb{I}[\mathcal{E}(y) = a^*]$, where $\mathcal{E}$ is a rule-based extractor that extracts the answer from a solution text and $\mathbb{I}$ is the indicator function with 0-1 binary output.

We use Group Relative Policy Optimization (GRPO) \citep{deepseek-math, deepseek-r1} to train the policy $\pi_\theta$. GRPO is a variant of Proximal Policy Optimization (PPO) \citep{ppo, gae} that uses group-based policy sampling to compute the advantage as the training signal, providing higher efficiency and stability without introducing additional value functions. For a given input $x$, GRPO generates a group of answers $y_1, \dots, y_G$ and optimizes the policy according to the estimated advantage derived from all responses in this group: 
\begin{equation}
    \label{eq:grpo}
    \begin{aligned}
        \mathcal{J}_{\text{GRPO}}(\theta)& = \mathbb{E}_{x \in \mathcal{D}, \{y_i\}_{i=1}^{G} \sim \pi_{\theta_{\text{old}}}(\cdot|x)} \\
        & \left[
        \frac{1}{G} \sum_{i=1}^{G} \left(
        \min\left(
        \frac{\pi_\theta(y_i|x)}{\pi_{\theta_{\text{old}}}(y_i|x)} A_i,
        \text{clip}\left(
        \frac{\pi_\theta(y_i|x)}{\pi_{\theta_{\text{old}}}(y_i|x)}, 1 - \epsilon, 1 + \epsilon
        \right) A_i
        \right)
        \right) - \beta \mathbb{D}_{\text{KL}}(\pi_\theta \| \pi_{\text{ref}})
        \right], 
    \end{aligned}
\end{equation}
where $\mathcal{D}$ is the dataset of math questions, $\pi_{\text{old}}$ is the old policy parameters for sampling, $\epsilon$ and $\beta$ are two hyper-parameters, and $\mathbb{D}_{\text{KL}}(\cdot \| \cdot)$ is the KL-divergence. The advantage $A_i$ of the generation $y_i$ is defined as 
\begin{equation}
    A_i = \frac{r_i - \text{mean}(r_1, \dots, r_G)}{\text{std}(r_1, \dots, r_G)},
\end{equation}
where $r_i$ is the reward on response $y_i$, which can be the correctness reward $r_c$ for standard GRPO. 

\subsection{Generative Verifiers}

LLM-based verifiers are widely used for LLM post-training and scaling LLM performance at inference time. Traditional verifiers are typically trained as discriminative classifiers to score the generated solution with additional network heads. Recently, generative verifiers \citep{generative-verifier,generative-rm,inference-time-rm} propose to train the verifier through token prediction, in the same way as classical text generation. Unlike discriminative verifiers, generative verifiers can synthesize verification rationales for improved explainability. 

A typical generative verifier will prompt the model to verify the solution $y$ for a given problem $x$. The model is required to answer \texttt{Yes} or \texttt{No} after a template $I$ like \texttt{Is the answer correct? (Yes/No)}. We also enable the model to generate intermediate rationales to justify its verification before outputting the final answer. This chain-of-thought (CoT) generation process has proven helpful for the final verification result. 

At inference time, to better measure the quality of each solution, we use the likelihood of the \texttt{Yes} token as the solution score. Let $y_v$ be the next token after the given template $I$, we have
\begin{equation}
    \label{eq:verifier-score}
    s(x, y) = P(y_v =\texttt{Yes}\mid x, y, I).
\end{equation}
In the original generative verifier work \cite{generative-verifier}, the model is trained through supervised fine-tuning (SFT) with well-constructed data. Our work considers adopting RL to train such a verifier model within the training of math reasoning. 

\begin{algorithm}[t]
\caption{GRPO with Self-Verification}
\label{alg:grpo-with-self-verification}
\begin{algorithmic}[1]
\STATE \textbf{Input:} Dataset $\mathcal{D}$, initial policy $\pi_{\theta}$, online buffer size $T_b$, group size $G$, total training steps $T$
\STATE \textbf{Initialize:} Policy-aligned buffer $\mathcal{B} \leftarrow \emptyset$
\FOR{$t=1, \dots, T$}
    \STATE Sample a data batch from the joint dataset $\mathcal{D} \cup \mathcal{B}$
    \FOR{input prompt $x$ in the batch}
        \STATE Generate a group of responses $\{y_i\}_{i=1}^{G}$ from $\pi_{\theta}$
        \IF{the input $x$ is from $\mathcal{D}$}
            \STATE $\triangleright$ Math reasoning problem
            \STATE Compute correctness rewards $r_c$ for each generation $y_1, \dots, y_G$
            \STATE $\mathcal{B} \leftarrow \mathcal{B} \cup \{ (x,y_1), \dots, (x,y_G) \}$
        \ELSE 
            \STATE $\triangleright$ Verification problem
            \STATE Compute verification reward $\hat{r}_v$ for each generation $y_1, \dots, y_G$ by \cref{eq:dynamic-verification-reward}
        \ENDIF
    \ENDFOR
    \STATE Compute the total reward $r = r_c + \hat{r}_v$ for each sample
    \STATE Update the model $\pi_\theta$ according to \cref{eq:grpo}
    \STATE Remove data samples from $\mathcal{B}$ that were collected before $t - T_b$ steps
\ENDFOR
\end{algorithmic}
\end{algorithm}

\section{Self-Verification Framework}

Since conventional test-time scaling methods often fail to generalize effectively for post-trained models, we introduce a \textit{self-verification} framework that enables both efficient multi-task reinforcement learning (RL) and robust test-time scaling. In \cref{section:rl}, we detail how answer generation and verification can be unified within a single RL process. Subsequently, in \cref{section:test-time-scaling}, we demonstrate how the trained model can leverage its verification capabilities to further enhance performance at inference time.

\subsection{RL with Self-Verification}
\label{section:rl}

We adopt the GRPO algorithm \citep{deepseek-math}, as described in \cref{sec:grpo}, to train large language models (LLMs) on math reasoning tasks. This approach utilizes a correctness reward $r_c(x, y)$ provided by a rule-based verifier. To extend the model's capabilities of verification, we need to introduce an additional reward function. Since we have access to the ground-truth correctness for each generated solution $y$, we can directly compare the model's predicted verification outcome with the ground truth:
\begin{equation}
    \label{eq:verification-reward}
    r_v(x, y) = \begin{cases}
        1, & \text{if } \mathbb{I}(y_v=\texttt{Yes}) = r_c(x, y), \\
        0, & \text{otherwise}.
    \end{cases}
\end{equation}
Here, the token $y_v$ is selected from \texttt{Yes} or \texttt{No}, representing the model's judgment on the verification task. Thus, $r_v$ serves as a direct training signal for the verifier. Notably, the data for the verification task depends not only on the original math problem but also on the model's generated response, so we use an online data buffer to store real-time model responses during training. 

\paragraph{Policy-Aligned Buffer.} 
We leverage solutions generated during math reasoning training as data to train the verifier. However, selecting appropriate verification data poses a challenge. Retaining all historically generated solutions would force the verifier to learn from potentially meaningless or unrefined solutions produced during early training stages. To address this issue, we emphasize that our verifier should primarily focus on solutions representative of the current model's capabilities. Therefore, we implement a policy-aligned buffer $\mathcal{B}$ that stores only the most recent solutions, mitigating the risks associated with off-policy data. Specifically, we maintain only solutions generated within the last $T_b$ training steps, ensuring the verification dataset is aligned with the evolving policy.

\paragraph{Dynamic Verification Reward.} 
When training the verifier through RL, we find that a simple binary reward as defined in \cref{eq:verification-reward} proves ineffective. As the model improves during training, we find that the correct solutions will soon dominate the verification dataset, creating an imbalanced data input. We want our verifier to focus on difficult verification problems during training. Hence, we introduce a dynamic verification reward that identifies challenging verification cases and dynamically adjusts rewards according to their difficulty. Leveraging the GRPO algorithm, we can obtain a group of generated responses $y_1, \dots, y_G$ for each problem $x$. Denoting $G_c$ and $G_i$ as the number of correct and incorrect generations in the group, we define the dynamic verification reward $\hat{r}_v$ as:
\begin{equation}
    \label{eq:dynamic-verification-reward}
    \hat{r}_v(x, y) = \begin{cases}
        \frac{2G_i}{G}, & \text{if } y_v = \texttt{Yes} \text{ and } r_c(x, y) = 1, \\
        \frac{2G_c}{G}, & \text{if } y_v = \texttt{No} \text{ and } r_c(x, y) = 0, \\
        0, & \text{otherwise}.
    \end{cases}
\end{equation}
This reward design provides higher incentives when the verifier successfully identifies a correct solution to a difficult problem or an incorrect solution for an easy problem. We maintain the expected reward for correct verification at $1$, ensuring it remains comparable in scale to the problem-solving reward $r_c$.

As detailed in Algorithm \ref{alg:grpo-with-self-verification}, our algorithm integrates self-verification into GRPO training by maintaining an online buffer of recent model generations. For each training step, we combine the original dataset and the buffer to sample problems. When processing dataset problems, we generate multiple solutions, compute their correctness rewards, and add them to the buffer. For problems from the buffer, we compute dynamic verification rewards based on the difficulty of verification. The final reward combines both correctness and verification components. This approach ensures the model learns to generate correct solutions while simultaneously developing verification capabilities.

\subsection{Self-Verification for Test-time Scaling}
\label{section:test-time-scaling}

At inference time, as our post-trained model possesses a strong built-in verifier, we can scale the model performance by allowing the model to verify its own solutions. Given a generation budget of $N$, we first sample $N$ candidate responses $y_1, \dots, y_N$ from the model. For each response, we compute a verification score $s(x, y_i)$ as defined in \cref{eq:verifier-score}. We then extract the set of unique answers $\mathcal{A} = \{\mathcal{E}(y_1), \dots, \mathcal{E}(y_N)\}$. We further aggregate the verification scores for each candidate answer to determine the final prediction $\hat{a}$:
\begin{equation}
    \label{eq:test-time-scaling}
    \hat{a} = \operatorname{argmax}_{a \in \mathcal{A}} \nolimits{\sum}_{i=1}^N \mathbb{I}[\mathcal{E}(y_i) = a] \left(1 + \alpha s(x, y_i)\right),
\end{equation}
where $\alpha$ is a hyperparameter that controls the scale of verification scores. When $\alpha = 0$, this reduces to standard majority voting; as $\alpha \rightarrow \infty$, the answer with the highest verification score dominates. By choosing a moderate value for $\alpha$, we can achieve a balance between consensus and confidence, leading to more robust and reliable predictions. A key advantage of this self-verification approach is its efficiency: it requires only a single LLM deployment at inference time, eliminating the need for external verifier models. This makes our framework highly practical and easily compatible with modern LLM inference engines such as vLLM \citep{vllm} and SGLang \citep{sglang}.

\section{Related Work}

\paragraph{RL for Reasoning.} The OpenAI o1 model series \citep{openai2024reasoning} has spurred a surge of research into the reasoning capabilities of large language models (LLMs). Early self-improvement methods \citep{star,vstar,llm-self-improvement} enable models to iteratively generate and refine their own reasoning paths by selecting high-quality solutions. With the emergence of DeepSeek-R1 \citep{deepseek-r1}, RL has become increasingly prominent as a means to directly incentivize solution correctness. Recent works including SimpleRL-Zoo \citep{simple-rl-zoo}, DeepScaleR \citep{deepscale-r}, Open R1 \citep{openr1}, and Light-R1 \citep{light-r1} demonstrate substantial improvements on reasoning tasks by applying RL to post-train pretrained models. Building on these advances, our work extends RL-based approaches to jointly train the model both for problem-solving and verification, enabling simultaneous improvement in these two tasks.

\paragraph{Test-time Scaling.} Numerous techniques have been developed to enhance LLM performance at inference time by increasing computational resources and promoting output diversity. Self-consistency decoding, for example, samples multiple solutions and aggregates them via majority voting \citep{self-consistency}, while some works also consider scaling the response length for more exhaustive reasoning paths \citep{s1}. Other approaches often leverage external reward models (RMs) to guide generation, including beam search \citep{deepmind-test-time-scaling-2,beam-search-reason}, Monte-Carlo tree search \citep{tree-of-thoughts, mcts-reason, mcts-alphazero}, multi-turn correction \citep{think-twice, score, self-correct, self-refine, reflexion}. For general pretrained models, process-based supervision—which rewards each correct reasoning step—has been shown to further improve performance \citep{process-outcome, process-evaluator, deepmind-test-time-scaling, deepmind-test-time-scaling-2}. However, our findings indicate that these process-based methods may be less effective for specifically post-trained models due to severe distributional shifts.

\paragraph{Verifier Training.} Robust verification mechanisms are crucial for reliable reasoning in LLMs. Early approaches typically train discriminative verifiers \citep{cobbe2021verifiers} by adding an auxiliary classification head to the model, with parameters learned from pairwise feedback \citep{instructgpt, rlaif} or ground-truth labels \citep{math-shepherd, rlhflow-prm}. Inspired by the LLM-as-a-judge paradigm \citep{llm-as-judge} used in the evaluation, recent work has explored generative verification, where the model produces verification results through text generation \citep{generative-verifier, generative-rm, inference-time-rm}. These generative verifiers can articulate chain-of-thought rationales \citep{chain-of-thought} when assessing solutions, thereby making the verification process more transparent and interpretable. Previous works usually train generative verifiers via supervised fine-tuning (SFT) on datasets containing high-quality verification responses, whereas a concurrent work \citep{rl-v} also considers using SFT to tune a verifier model within varying RL training algorithms. To our knowledge, our work is the first to propose a unified RL framework that unifies problem-solving and verification.

\section{Experiments}

\begin{table}[t]
    \centering
    \caption{Average \textbf{greedy-decoding scores} of different models on math reasoning benchmarks after post-training. The best scores from each model series are highlighted in bold. For AIME24, AIME25, and AMC23, we report the average scores over 10 samples for each problem. }
    \label{tab:post-training-results}
    \begin{tabular}{lccccc}
    \toprule
    Model & MATH500 & \makecell{AIME24\\(avg@10)} & \makecell{AIME25\\(avg@10)} & \makecell{AMC23\\(avg@10)} & \makecell{Olympiad\\Bench} \\
    \midrule
    \multicolumn{6}{l}{\textit{Model Series: Qwen2.5-Math-7B}} \\
    \addlinespace[2pt]
    \makecell[l]{Self-Verification-Qwen-7B (Ours)\\(Problem-solving + verification)} & \textbf{83.60} & 20.00 & \textbf{16.67} & 63.75 & \textbf{34.81} \\
    \addlinespace[1pt]
    \makecell[l]{Qwen2.5-Math-7B\\(Base model)} & 62.00 & 14.67 & 5.00 & 45.25 & 17.63 \\
    \addlinespace[1pt]
    \makecell[l]{GRPO-Qwen-7B\\(Problem-solving Only)} & 81.40 & 19.67 & 15.67 & \textbf{65.50} & 32.89 \\
    \addlinespace[1pt]
    \makecell[l]{SimpleRL-Qwen-Math-7B (\citep{simple-rl-zoo})} & 80.80 & \textbf{23.33} & 10.00 & 63.75 & 32.15 \\
    \midrule
    \multicolumn{6}{l}{\textit{Model Series: DeepSeek-R1-Distill-Qwen-1.5B}} \\
    \addlinespace[2pt]
    \makecell[l]{Self-Verification-R1-1.5B (Ours)\\(Problem-solving + verification)} & \textbf{87.00} & \textbf{43.00} & \textbf{31.33} & \textbf{77.50} & \textbf{44.30} \\
    \addlinespace[1pt]
    \makecell[l]{R1-Distill-Qwen-1.5B\\(Base model)} & 80.00 & 24.33 & 25.00 & 64.25 & 32.89 \\
    \addlinespace[1pt]
    \makecell[l]{GRPO-R1-1.5B\\(Problem-solving only)} & \textbf{87.00} & 37.67 & 26.67 & 72.50 & 40.74 \\
    \addlinespace[1pt]
    \makecell[l]{DeepScaleR-1.5B-Preview (\citep{deepscale-r})} & 83.00 & 37.00 & 31.00 & 77.25 & 43.56 \\
    \bottomrule
    \end{tabular}
\end{table}

In this section, we aim to examine the effectiveness and characteristics of our self-verification framework. We compare our methods with standard RL and test-time baselines to answer the following questions: (1) How does imposing self-verification on the RL process compare to standard RL for problem-solving? (2) Can the learned model verify its own solutions better than external verifiers? (3) How does self-verification help with test-time scaling? (4) Is the self-verification process efficient enough for post-training and test-time scaling?

\paragraph{Models and Training.} 
In our experiments, we primarily use two pretrained models for RL training with self-verification, which are Qwen2.5-Math-7B \citep{qwen25-math} and DeepSeek-R1-Distill-Qwen-1.5B \citep{deepseek-r1}. The Qwen2.5-Math-7B model is a 7B-parameter model pretrained in massive math data with a relatively short generation length, while DeepSeek-R1-Distill-Qwen-1.5B is a distilled version of the powerful DeepSeek-R1 model with the ability of generating long CoT responses. We use the popular \texttt{verl} framework \citep{verl} as the code base for RL training, where the maximal context length for Qwen2.5-Math-7B is 4k and for DeepSeek-R1-Distill-Qwen-1.5B is 16k. We name the trained models as Self-Verification-Qwen-7B and Self-Verification-R1-1.5B, respectively. We provide our training details in Appendix~\ref{app:post-training}.

\begin{table}[t]
    \centering
    \caption{The performance of different models on verifying MATH500 solutions generated by the Self-Verification-Qwen-7B model. We highlight the best scores from the open-source models in bold. }
    \label{tab:verification-result-qwen}
    \begin{tabular}{llcc}
        \toprule
        \textbf{Category} & \textbf{Method} & \textbf{Accuracy} & \textbf{F1 Score} \\
        \midrule
        \multirow{3}{*}{\makecell{Open-source Models\\(\textasciitilde 7B)}} 
            & Self-Verification-Qwen-7B (Ours) & \textbf{87.20} & \textbf{92.83} \\
            & Qwen2.5-Math-7B (Base model) & 73.20 & 84.93 \\
            & Llama-3.1-8B-Instruct & 67.00 & 78.20 \\
        \midrule
        \multirow{3}{*}{Proprietary Models} 
            & GPT-4o & 85.20 & 91.57 \\
            & Claude-3.7-Sonnet & 90.20 & 94.46 \\
            & DeepSeek-v3 & 89.00 & 93.73 \\
        \bottomrule
    \end{tabular}
\end{table}

\begin{table}[t]
    \centering
    \caption{The performance of different models on verifying AIME24 solutions generated by the Self-Verification-R1-1.5B model. We highlight the best scores from the open-source models in bold.}
    \label{tab:verification-result-r1}
    \begin{tabular}{llcc}
        \toprule
        \textbf{Category} & \textbf{Method} & \textbf{Accuracy} & \textbf{F1 Score} \\
        \midrule
        \multirow{4}{*}{\makecell{Open-source Models\\(1.5B \& \textasciitilde 7B)}} 
            & Self-Verification-R1-1.5B (Ours) & \textbf{56.67} & \textbf{67.72} \\
            & R1-Distill-Qwen-1.5B (Base model) & 38.00 & 49.46 \\
            & R1-Distill-Qwen-7B & 46.00 & 59.50 \\
            & Llama-3.1-8B-Instruct & 55.67 & 45.71 \\
        \midrule
        \multirow{3}{*}{Proprietary Models} 
            & GPT-4o & 59.33 & 65.54 \\
            & Claude-3.7-Sonnet & 64.33 & 71.16 \\
            & DeepSeek-v3 & 57.67 & 66.67 \\
        \bottomrule
    \end{tabular}
\end{table}

\begin{table}[t]
    \centering
    \caption{Average \textbf{test-time scaling} scores of different methods on various math reasoning benchmarks. All the test-time scaling methods have a budget of 16 samples for each problem. The best scores from each model series are highlighted in bold. For AIME24, AIME25, and AMC23, we report the average scores over 10 samples for each problem. }
    \label{tab:test-time-scaling}
    \begin{tabular}{lccccc}
        \toprule
        Method@16 & MATH500 & \makecell{AIME24\\(avg@10)} & \makecell{AIME25\\(avg@10)} & \makecell{AMC23\\(avg@10)} & \makecell{Olympiad\\Bench} \\
        \midrule
        \multicolumn{6}{l}{\textit{Model Series: Qwen2.5-Math-7B}} \\
        \addlinespace[2pt]
            Self-Verification (Ours) & \textbf{87.20} & \textbf{26.67} & 19.00 & \textbf{73.25} & \textbf{39.70} \\
            Self-Consistency & 86.00 & 23.67 & \textbf{21.67} & 71.25 & 39.11 \\
            Best-of-N        & 85.00 & 23.33 & 16.67 & 64.25 & 37.92 \\
            Beam Search      & 82.80 & 21.00 & 15.00 & 67.25 & 35.71 \\
            DVTS             & 82.80 & 21.67 & 20.33 & 67.50 & 35.85 \\
        \midrule
        \multicolumn{6}{l}{\textit{Model Series: DeepSeek-R1-Distill-Qwen-1.5B}} \\
        \addlinespace[2pt]
            Self-Verification (Ours) & \textbf{93.60} & \textbf{57.67} & \textbf{37.67} & \textbf{92.00} & \textbf{50.96} \\
            Self-Consistency & 91.00 & 52.00 & 36.67 & 87.50 & 47.56 \\
            Best-of-N        & 86.00 & 46.33 & 33.67 & 83.75 & 43.41 \\
            Beam Search      & 88.40 & 42.67 & 33.00 & 85.25 & 44.59  \\
            DVTS             & 90.80 & 45.33 & 32.33 & 82.50 & 45.18 \\
        \bottomrule
    \end{tabular}
\end{table}

\paragraph{Data and Benchmarks.} 
We target our self-verification framework on mathematical reasoning tasks. Considering previous popular implementations of GRPO on math reasoning \citep{simple-rl-zoo, deepscale-r}, we use the level 3-5 data from the math training dataset \cite{math-dataset} to train the Qwen2.5-Math-7B model and a combined math reasoning dataset sorted by DeepScaleR \citep{deepscale-r} to train the DeepSeek-R1-Distill-Qwen-1.5B model. For the evaluation benchmarks, we adopt the MATH500 dataset \cite{prm-openai}, AIME 2024 and 2025 problems, AMC 2023 problems, and OlympiadBench \citep{olympiadbench}, which are commonly used benchmarks for evaluating math reasoning models \cite{qwen25-math}. We enable a context length of 4k for methods based on Qwen-2.5-Math-7B and a context length of 16k for methods based on DeepSeek-R1-Distill-Qwen-1.5B.  Unless stated otherwise, we use the accuracy of the generated answers as the score shown in the experimental results, which is automatically verified by the \texttt{math-verify} library from HuggingFace\footnote{\url{https://github.com/huggingface/Math-Verify}}. As the evaluation data is relatively scarce in AIME24, AIME25, and AMC23, we repeat each problem 10 times and use the average accuracy as the result. 

\paragraph{Baselines.} Our baselines include standard RL methods and test-time scaling baselines. For the RL algorithms, we compare the standard GRPO version without self-verification training and name them GRPO-Qwen-7B and GRPO-R1-1.5B, respectively. To compare the validity of our RL process with prior works, we also include SimpleRL-Qwen-Math-7B model \citep{simple-rl-zoo} and DeepScaleR-1.5B-Preview model \citep{deepscale-r} as our baselines, which are also tuned based on Qwen2.5-Math-7B and DeepSeek-R1-Distill-Qwen-1.5B. For the test-time scaling baselines, we include the following methods: 
\begin{itemize}
    \item \textbf{Self-consistency} \citep{self-consistency} uses multiple CoT solutions to improve the accuracy of the final answer through majority voting.
    \item \textbf{Best-of-N} \citep{deepmind-test-time-scaling-2} adopts the sample with the highest score from $N$ generated responses.
    \item \textbf{Process-based test-time scaling} methods adopt a process-based reward model (PRM) \citep{prm-openai,math-shepherd} to further supervise the generation in each step with various strategies like \textbf{beam search} \citep{deepmind-test-time-scaling,deepmind-test-time-scaling-2} and \textbf{Diverse Verifier Tree Search (DVTS)} \citep{hf-test-time-scaling}. 
\end{itemize}
We can split the baselines into two categories. The self-consistency method is a simple ensemble method without using other models, while other baselines including best-of-N, beam search, and DVTS require external reward models to guide generations. For the choice of models, we follow \citet{hf-test-time-scaling} to use the RLHFlow \texttt{Llama3.1-8B-PRM-Deepseek-Data} PRM with 8B parameters \citep{math-shepherd, rlhflow-prm}, which is the best PRM revealed in their work with similar size to our Qwen-based 7B verifier model. 

\subsection{Post-training Model Performance}

\paragraph{Performance on Problem-Solving} 
We begin by examining the impact of incorporating self-verification during post-training on the greedy-decoding performance of the model, without employing any additional test-time techniques. To this end, we evaluate our self-verification models across several math reasoning benchmarks, as summarized in \cref{tab:post-training-results}. Both Self-Verification-Qwen-7B and Self-Verification-R1-1.5B consistently achieve strong results, outperforming their respective base models by a significant margin due to the reinforcement learning process. Notably, integrating self-verification into RL yields even higher scores than the standard GRPO models, suggesting that self-verification intrinsically enhances the model's problem-solving abilities. We speculate that this synergy arises because the verification task encourages the model to develop a deeper understanding of the problem's logical structure, which in turn enhances its problem-solving capabilities. Although no new input is introduced, verification training elicits knowledge and reasoning patterns that generalize back to the generation task. 

\paragraph{Performance on Verifying Solutions} 
We further evaluate the effectiveness of our self-verification models in verifying their own generated solutions. Specifically, we prompt various LLMs to verify the solution for a given problem, extract their binary verification answer (\texttt{Yes} or \texttt{No}), and compare it against the ground-truth correctness. As presented in \cref{tab:verification-result-qwen} and \cref{tab:verification-result-r1}, we conduct this generative verification procedure with multiple LLMs, including open-source models of comparable size to ours as well as proprietary models accessed via APIs. On the MATH500 and AIME24 benchmarks, our self-verification models demonstrate substantial improvements in both accuracy and F1 score over similarly sized open-source baselines. Notably, our models also achieve performance on par with leading commercial systems such as GPT-4o \citep{gpt-4o}, Claude-3.7-Sonnet \citep{claude-37-sonnet}, and DeepSeek-v3 \citep{deepseek-v3}. Our model can even surpass GPT-4o in verifying their own responses despite having significantly fewer parameters. We also observe that verifying AIME24 solutions poses a greater challenge for all models, likely due to the increased complexity of the problems. 

\begin{figure}[h]
    \centering
    \begin{minipage}[t]{0.49\textwidth}
        \centering
        \includegraphics[width=0.9\linewidth]{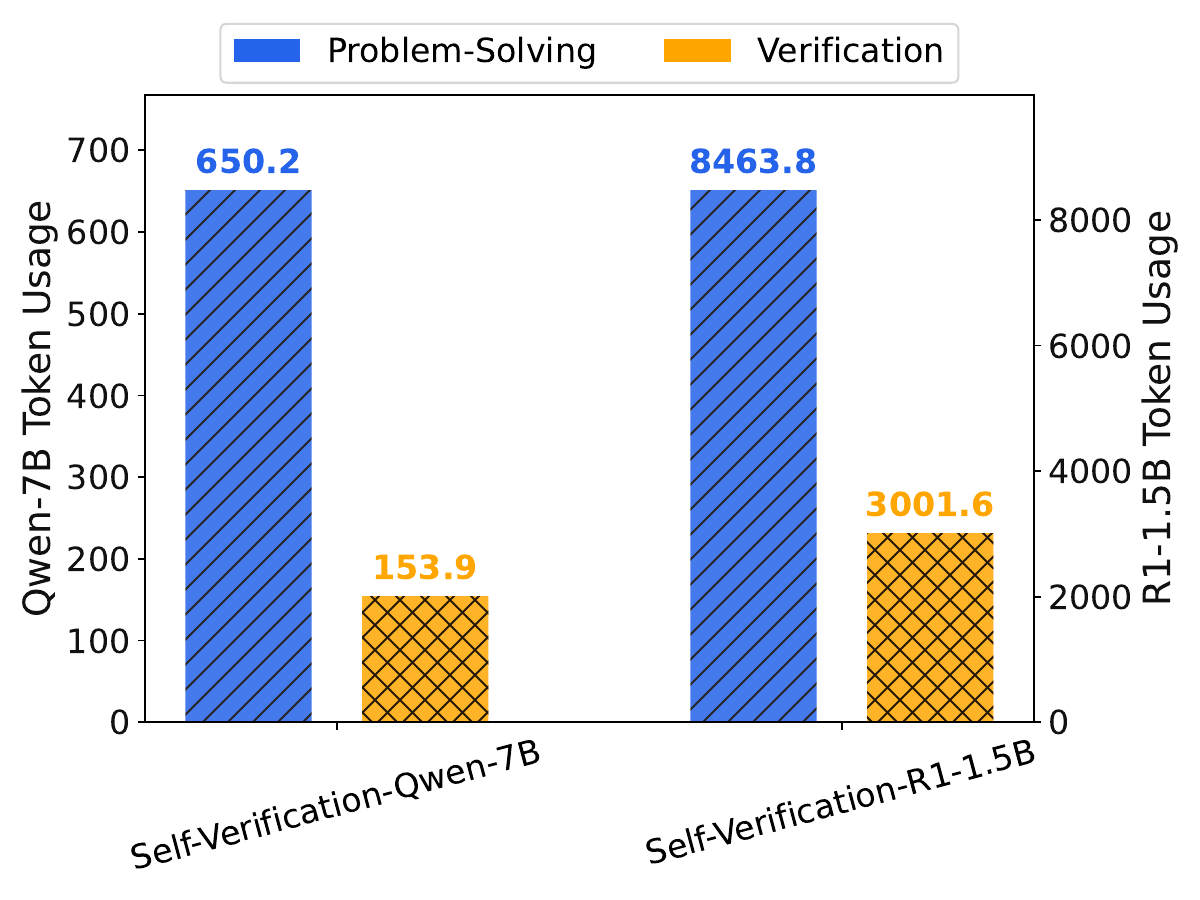}
        \caption{Token usage comparison between problem-solving and verification tasks.}
        \label{fig:token-usages}
    \end{minipage}
    \hfill
    \begin{minipage}[t]{0.49\textwidth}
        \centering
        \includegraphics[width=0.9\linewidth]{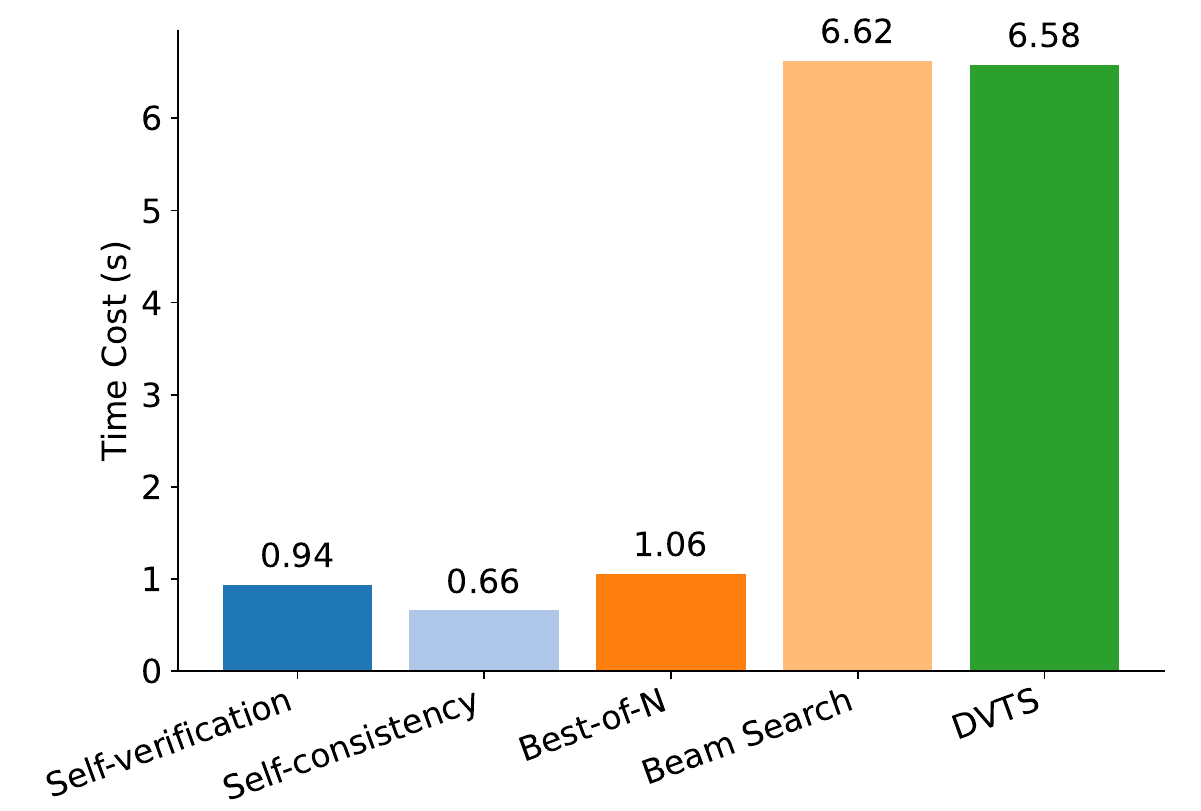}
        \caption{Average time cost of different test-time scaling methods per problem from MATH500.}
        \label{fig:time-cost}
    \end{minipage}
\end{figure}

\subsection{Self-verification on Test-time Scaling}

As stated in \cref{section:test-time-scaling}, our trained self-verification model can scale its performance by verifying its generated answers. We use Self-Verification-Qwen-7B and Self-Verification-R1-1.5B to generate solutions with a given sample budget and further adopt different test-time scaling methods to evaluate the performance of the final answer. In \cref{fig:scaling-figure}, we show the scaling performance of different methods when sampling from Self-Verification-Qwen-7B on MATH500 and Self-Verification-R1-1.5B on AIME24. The results show that test-time scaling with self-verification yields the best performance among all methods. Maybe surprisingly, we find that the simple self-consistency method also presents competitive performance by majority voting. In contrast, methods based on external verifiers like best-of-N, beam search, and DVTS perform worse than self-consistency. This observation proves our hypothesis that external verifiers trained on specific data do not generalize to verifying our post-trained models. Process-based methods like beam search and DVTS adopt a step-wise generation, which may further amplify the distribution shift during the generation process. In \cref{tab:test-time-scaling}, we show the detailed results on more math reasoning benchmarks with a generation budget of 16, which is a common number of generations balancing between the quality and the efficiency. We see that our self-verification models can consistently outperform all other methods on most tasks. 

\begin{table}[t]
    \centering
    \caption{Ablation of the aggregation coefficient $\alpha$ on MATH500 with a 16-sample budget. Accuracy is reported for different $\alpha$ settings when selecting answers using verification probabilities.}
    \label{tab:alpha-ablation}
    \begin{minipage}[t]{0.45\textwidth}
        \centering
        \begin{tabular}{cc}
            \toprule
            \multicolumn{2}{c}{\textbf{Self-Verification-Qwen-7B@16}} \\
            \midrule
            $\alpha$ & \textbf{Accuracy} \\
            \midrule
            0.3 & 86.60 \\
            1.0 & \textbf{87.20} \\
            3.0 & 87.10 \\
            5.0 & 87.10 \\
            \bottomrule
        \end{tabular}
    \end{minipage}
    \hfill
    \begin{minipage}[t]{0.45\textwidth}
        \centering
        \begin{tabular}{cc}
            \toprule
            \multicolumn{2}{c}{\textbf{Self-Verification-R1-1.5B@16}} \\
            \midrule
            $\alpha$ & \textbf{Accuracy} \\
            \midrule
            0.03 & 92.60 \\
            0.10 & \textbf{93.60} \\
            0.30 & 92.40 \\
            1.00 & 91.00 \\
            \bottomrule
        \end{tabular}
    \end{minipage}
\end{table}

\subsection{Analysis on the Efficiency of Self-Verification}
In the post-training stage, our framework is comparable in training cost to other RL methods. As the verification process is expected to be less difficult than problem-solving, we find that the average tokens the model spends on verification is actually lower for both short-context Qwen 7B model and long-context R1 distilled model. As shown in \cref{fig:token-usages}, the average token usage in the verification task is 24\% of the problem-solving task for Self-Verification-Qwen-7B and 35\% for Self-Verification-R1-1.5B. The lower token usage makes our test-time scaling method more efficient since the verification process takes much less time than the problem-solving process. \cref{fig:time-cost} illustrates the average time cost per problem on the MATH500 benchmark for various test-time scaling methods, evaluated on a fixed hardware platform. Unlike approaches that rely on external Reward Models (RMs), our unified model architecture requires serving only a single LLM, which reduces overhead and improves efficiency. Our self-verification method is faster than best-of-N sampling and demonstrates significant efficiency gains over process-based methods like beam search and DVTS. Although its time cost is slightly higher than simple majority voting, this represents a favorable trade-off for improved performance. We report the implementation details of our inference-time scaling in Appendix~\ref{app:test-time-scaling}. 

\paragraph{Ablation Studies} We investigate how the mechanisms and hyperparameters proposed in our methodology affect the post-training and test-time performance. First, we examine the test-time aggregation coefficient $\alpha$, which balances majority voting with verifier confidence. We show the MATH500 performance of our two self-verification models in \cref{tab:alpha-ablation}. The Self-Verification-Qwen-7B model peaks at $\alpha = 1.0$ while not showing significant performance degradation across different $\alpha$ settings. In contrast, Self-Verification-R1-1.5B is more sensitive to this hyperparameter, preferring a smaller $\alpha=0.1$, likely due to its small model size that limits the verification capability. We also report the ablation results on our two post-training mechanisms, including the policy-aligned buffer and the dynamic verification reward, in Appendix~\ref{app:ablation-studies}.

\section{Conclusion and Limitations}
\label{section:conclusion}
In this paper, we introduced a novel self-verification framework for enhancing the mathematical reasoning of LLMs. Our approach integrates verification capabilities directly into the problem-solving model through a specialized RL process, enabling models to evaluate their own solutions in a generative verification process. Through extensive experiments across multiple mathematical benchmarks, we demonstrated that our self-verification models not only achieve superior performance in problem-solving tasks but also excel at verifying solution correctness. Our self-verification framework addresses the distribution shift issue that arises during test-time scaling with external verifiers, achieving better test-time scaling performance compared to other baselines. 

Our work has several limitations. First, the self-verification framework is primarily tailored for mathematical reasoning tasks, and its direct applicability to other domains such as code generation and agentic tasks, remains to be tested. Adapting self-verification to these areas may require additional task-specific designs, integration with external tools, or new verification strategies. Second, our test-time scaling by self-verification focuses on aggregating answers from multiple generations but does not consider the scale of response length. A multi-turn response generation between problem-solving and verification may further improve inference-time performance, which is a promising direction to extend our framework. We discuss the broader impact of our work in Appendix~\ref{app:broader-impact}.

\begin{ack}
Bo An is supported by the National Research Foundation Singapore and DSO National Laboratories under the AI Singapore Programme (AISGAward No: AISG4-GC-2023-009-1B).
\end{ack}

\bibliography{reference}
\bibliographystyle{unsrtnat}


\newpage
\appendix

\section{Broader Impact}
\label{app:broader-impact}

Our self-verification framework for LLMs has several potential societal impacts, both positive and negative, that warrant discussion.

\paragraph{Positive Impacts} The primary beneficial impact of our work is the improvement of LLM reliability in reasoning tasks. By enabling models to verify their own outputs, we can contribute to reducing hallucinations and factual errors in AI systems, which is crucial for applications in education, scientific research, and decision support. The self-verification approach also improves computational efficiency compared to methods requiring external verifiers, potentially reducing energy consumption and computational costs associated with deploying reasoning systems at scale. Additionally, our unified model approach demonstrates that complementary capabilities (reasoning and verification) can be trained together, which may inspire similar approaches in other domains requiring accuracy and trustworthiness.

\paragraph{Risks and Limitations} Despite the improvements in verification capabilities, our approach does not eliminate the risk of incorrect outputs or the model confidently asserting wrong answers. There is a potential concern that users might place excessive trust in self-verified outputs, believing them to be more reliable than they actually are, particularly in high-stakes domains like healthcare or financial analysis. The verification capability is also currently limited to mathematical reasoning tasks, and the generalizability to other domains remains uncertain. Furthermore, the increased capabilities could potentially be misused in generating deceptive content that appears more legitimate due to self-verification signals.

\paragraph{Safeguards and Mitigations} To address these concerns, we recommend several safeguards: (1) Clear communication to users about the limitations of self-verification, including transparency about error rates; (2) Complementary use of human verification for critical applications; (3) Continued research into detecting when models are operating outside their reliable domains; and (4) Development of benchmark tests to evaluate verification capabilities across diverse problem types and difficulty levels. We have also published our methodology transparently to enable further research into both the capabilities and limitations of self-verification approaches.

The overall aim of our work is to contribute to the development of more reliable and trustworthy AI systems, with self-verification representing one component in the broader ecosystem of techniques needed for responsible AI deployment.

\section{Technical Details on Post-training}
\label{app:post-training}

\begin{table}[t]
    \centering
    \caption{GRPO hyperparameters on the post-training stage with self-verification}
    \label{tab:post-training-hyperparams}
    \begin{tabular}{lll}
        \toprule
        \textbf{Hyperparameter} & \textbf{Self-Verification-Qwen-7B} & \textbf{Self-Verification-R1-1.5B} \\
        \midrule
        \multicolumn{3}{l}{\textit{Model and Data}} \\
        Base Model & Qwen2.5-Math-7B & DeepSeek-R1-Distill-1.5B \\
        Training Data & MATH-level3to5 \citep{simple-rl-zoo} & DeepScaleR \citep{deepscale-r} \\
        Max Prompt Length & 2048 & 2048 \\
        Max Response Length & 2048 & 14336 \\
        \midrule
        \multicolumn{3}{l}{\textit{RL Hyperparameters}} \\
        Train Batch Size & 128 & 128 \\
        PPO Mini-batch Size & 64 & 64 \\
        Learning Rate & 1e-6 & 1e-6 \\
        KL Loss Coefficient & 0.001 & 0.001 \\
        Entropy Coefficient & 0.001 & 0.005 \\
        Adaptive Entropy & False & True \\
        Target Entropy & N/A & 0.2 \\
        Entropy Coeff Delta & N/A & 0.0001 \\
        Max/Min Entropy Coeff & N/A & 0.005/0 \\
        \midrule
        \multicolumn{3}{l}{\textit{Rollout Settings}} \\
        Engine & vLLM \citep{vllm} & vLLM \citep{vllm} \\
        GPU Memory Utilization & 0.8 & 0.8 \\
        GRPO Group Size & 8 & 8 \\
        Temperature & 0.6 & 0.6 \\
        \midrule
        \multicolumn{3}{l}{\textit{Training Schedule}} \\
        Total Training Steps & 1000 & 2000 \\
        Rejection Sampling & $\checkmark$ & $\checkmark$ \\
        \midrule
        \multicolumn{3}{l}{\textit{Online Data Buffer}} \\
        Data Buffer Size & 5000 & 40000 \\
        Save Frequency $T_b$ & 60 & 60 \\
        \bottomrule
    \end{tabular}
\end{table}

In this section, we provide comprehensive details about our post-training process with self-verification. We use the popular verl framework\footnote{\url{https://github.com/volcengine/verl}}, which is the open-source version of \citet{verl}, as our RL training framework. For the training resources of our models, we use two nodes of 8 NVIDIA GPUs with 80GB memory each. The detailed training configurations and hyperparameters are listed in \cref{tab:post-training-hyperparams}.

\paragraph{Model and Data Configuration} For Self-Verification-Qwen-7B, we use Qwen2.5-Math-7B as our base model and train on the MATH training data \citep{math-dataset} with level 3 to 5 curated by SimpleRL-Zoo \citep{simple-rl-zoo}. The model operates with a maximum prompt length of 2048 tokens and a maximum response length of 2048 tokens. For Self-Verification-R1-1.5B, we use DeepSeek-R1-Distill-1.5B as the base model and train on the comprehensive dataset from DeepScaleR \citep{deepscale-r}, which is a collection from multiple math reasoning data sources. While keeping the same prompt length limit, we extend the maximum response length to 14336 tokens to accommodate the model's capability for generating longer chain-of-thought reasoning. Different from most math RL implementations, we use a longer prompt length of 2048 as we expect more input tokens for the verification task. For the Self-Verification-R1-1.5B model, we remove the thinking content from the original response, wrapped by the \texttt{<think>...</think>} tags, to reduce the input prompt length, which also follows the implementation of DeepSeek-R1 \citep{deepseek-r1} when building multi-turn chat conversation.

\paragraph{RL Settings} We adopt the GRPO algorithm for both models with the same batch size: a training batch size of 128 and a PPO mini-batch size of 64. Other training-related configurations are similar to the original implementation in verl. The learning rate is set to 1e-6 for both models. For the KL loss coefficient, we use 0.001 to maintain a balance between exploration and policy improvement. The entropy coefficient is set differently: 0.001 for Qwen-7B and 0.005 for R1-1.5B. Notably, we follow \citet{skywork-or1-2025} to enable adaptive entropy for R1-1.5B with a target entropy of 0.2, allowing the entropy coefficient to adjust between 0 and 0.005 per step, as we find that a fixed entropy coefficient for the DeepSeek-R1-Distill-1.5B model results in an extremely unstable training process. We leverage the vLLM engine \citep{vllm} for efficient inference during training, setting the GPU memory utilization to 0.8. For both models, we use a group size of 8 for GRPO and a temperature of 0.6 for sampling. The training runs for 1500 steps for Qwen-7B and 2000 steps for R1-1.5B, with rejection sampling enabled to filter out invalid generations.

\paragraph{Additional Rewards for Qwen-7B Model} When using the Qwen2.5-Math-7B model for RL, we find that the generated output of this base model is unstable, with useless and unverified code text, potentially due to its special pretraining data. Surprisingly, we find that the unexpected code snippets in the generated output do not significantly affect the RL performance in standard RL process. However, this meaningless code generation can be harmful for verification tasks. As a result, we add an additional code generation penalty of $-0.5$ for all generations containing code output. Additionally, we also add a short response penalty of $-0.5$ for the verification task when the model only outputs the final verification result without CoT. These rewards are specific to the Qwen 7B model. We believe that these auxiliary rewards have a minor effect on our main methodology since the model quickly convergess to these rewards in a few RL training steps. 

\paragraph{Online Buffer Management} The online buffer is crucial for our self-verification framework. For Qwen-7B, we maintain a buffer size of 5000 samples, while for R1-1.5B, we expand it to 40000 samples to accommodate the larger dataset and longer responses. Both models update their buffers every 60 training steps ($T_b$) to ensure the verification data stays aligned with the current policy.

\begin{table}[t]
    \centering
    \caption{Hyperparameters for test-time scaling with self-verification and baselines}
    \label{tab:test-time-scaling-config}
    \begin{tabular}{lll}
        \toprule
        \textbf{Parameter Category} & \textbf{Self-Verification-Qwen-7B} & \textbf{Self-Verification-R1-1.5B} \\
        \midrule
        \multicolumn{3}{l}{\textit{Generation Settings}} \\
        Number of Samples & 16 (unless specified) & 16 (unless specified) \\
        Temperature & 0.6 & 0.6 \\
        Top-$p$ & 1.0 & 1.0 \\
        Max Tokens & 2048 & 14336 \\
        \midrule
        \multicolumn{3}{l}{\textit{Self-verification Configuration}} \\
        Verifier Max Tokens & 2048 & 14336 \\
        Verifier Temperature & 0.0 & 0.0 \\
        Verification Scale Coefficient $\alpha$ & 1.0 & 0.1 \\
        \midrule
        \multicolumn{3}{l}{\textit{Process-based Methods (Beam Search and DVTS) Configuration}} \\
        Beam Width & \makecell[l]{4, or\\2 when number of samples is 2} & \makecell[l]{4, or\\2 when number of samples is 2} \\
        Step Delimiter & two linebreaks & two linebreaks \\
        Max Search Steps & 40 & 40 \\
        \bottomrule
    \end{tabular}
\end{table}

\begin{figure}[t]
    \centering
    \begin{subfigure}[t]{0.48\textwidth}
        \centering
        \includegraphics[width=\linewidth]{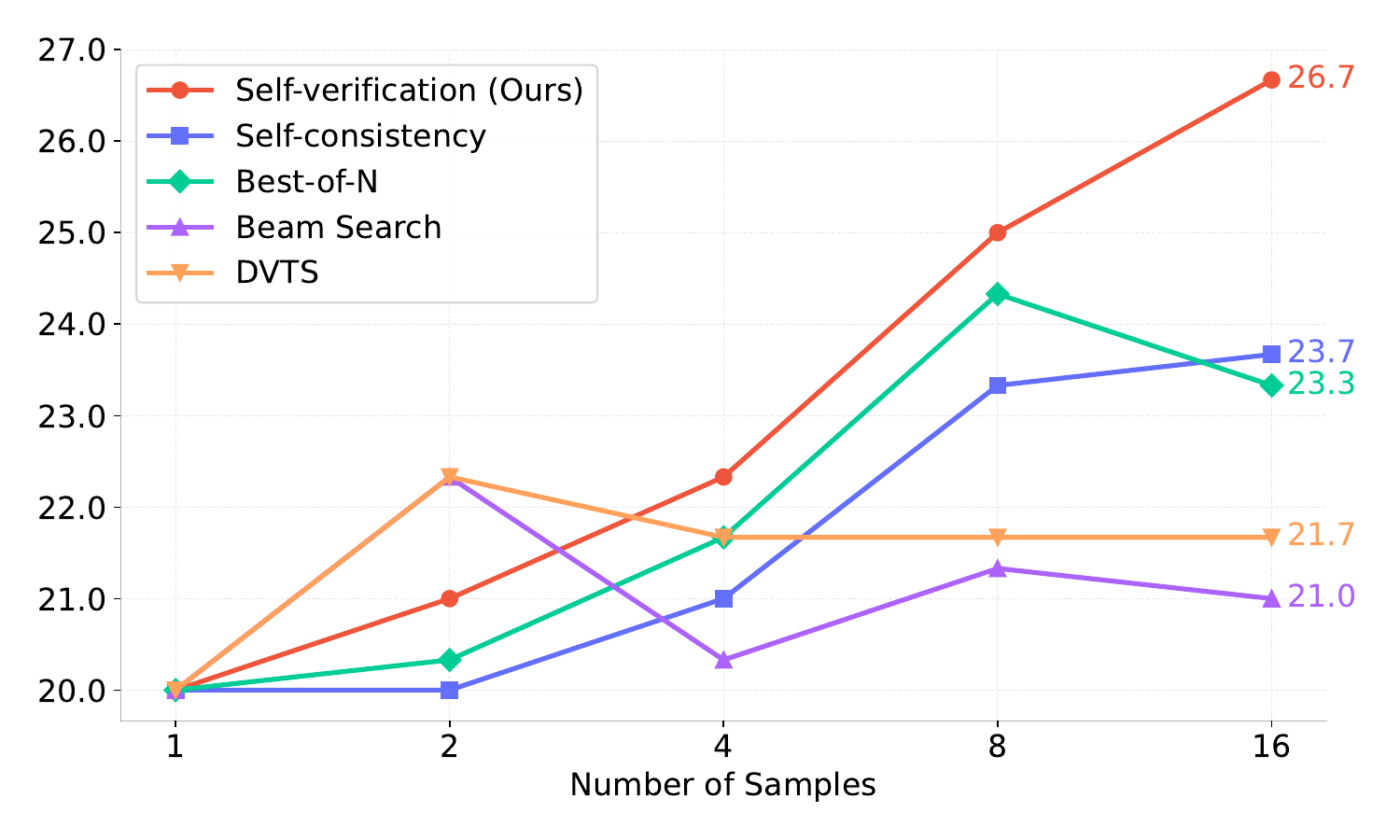}
        \caption{AIME24 Score (avg@10)}
        \label{fig:qwen-7b-aime24}
    \end{subfigure}
    \hfill
    \begin{subfigure}[t]{0.48\textwidth}
        \centering
        \includegraphics[width=\linewidth]{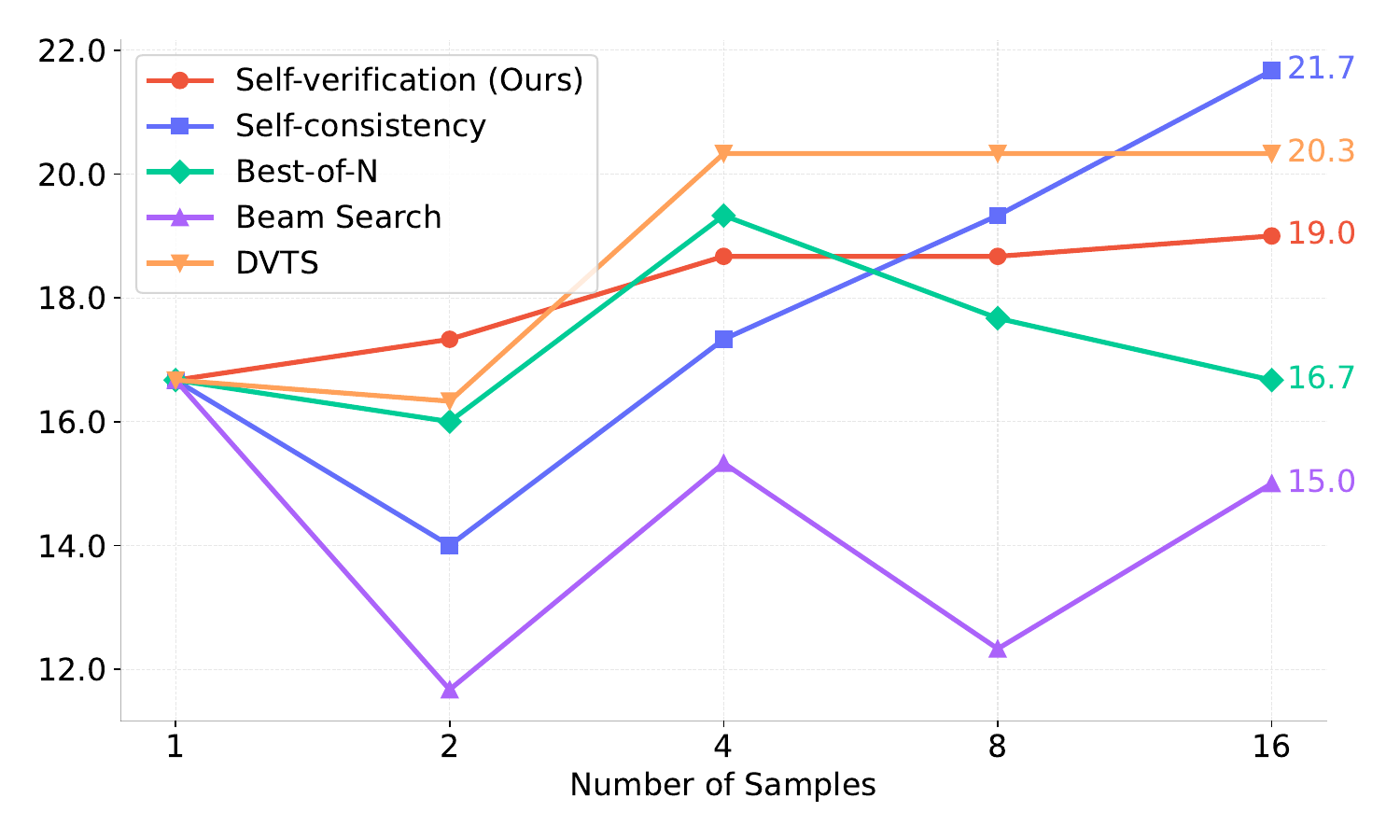}
        \caption{AIME25 Score (avg@10)}
        \label{fig:qwen-7b-aime25}
    \end{subfigure}
    
    \begin{subfigure}[t]{0.48\textwidth}
        \centering
        \includegraphics[width=\linewidth]{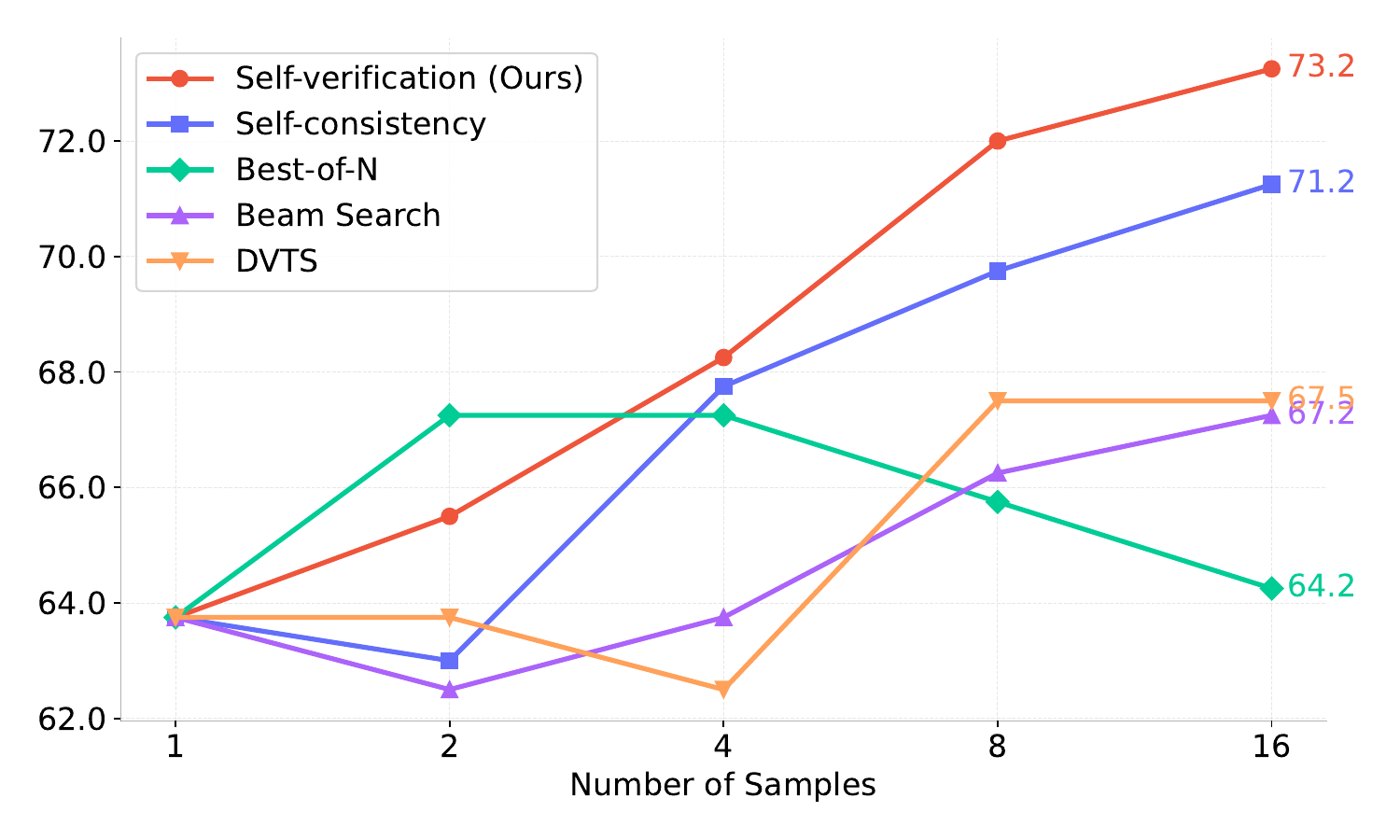}
        \caption{AMC23 Score (avg@10)}
        \label{fig:qwen-7b-amc23}
    \end{subfigure}
    \hfill
    \begin{subfigure}[t]{0.48\textwidth}
        \centering
        \includegraphics[width=\linewidth]{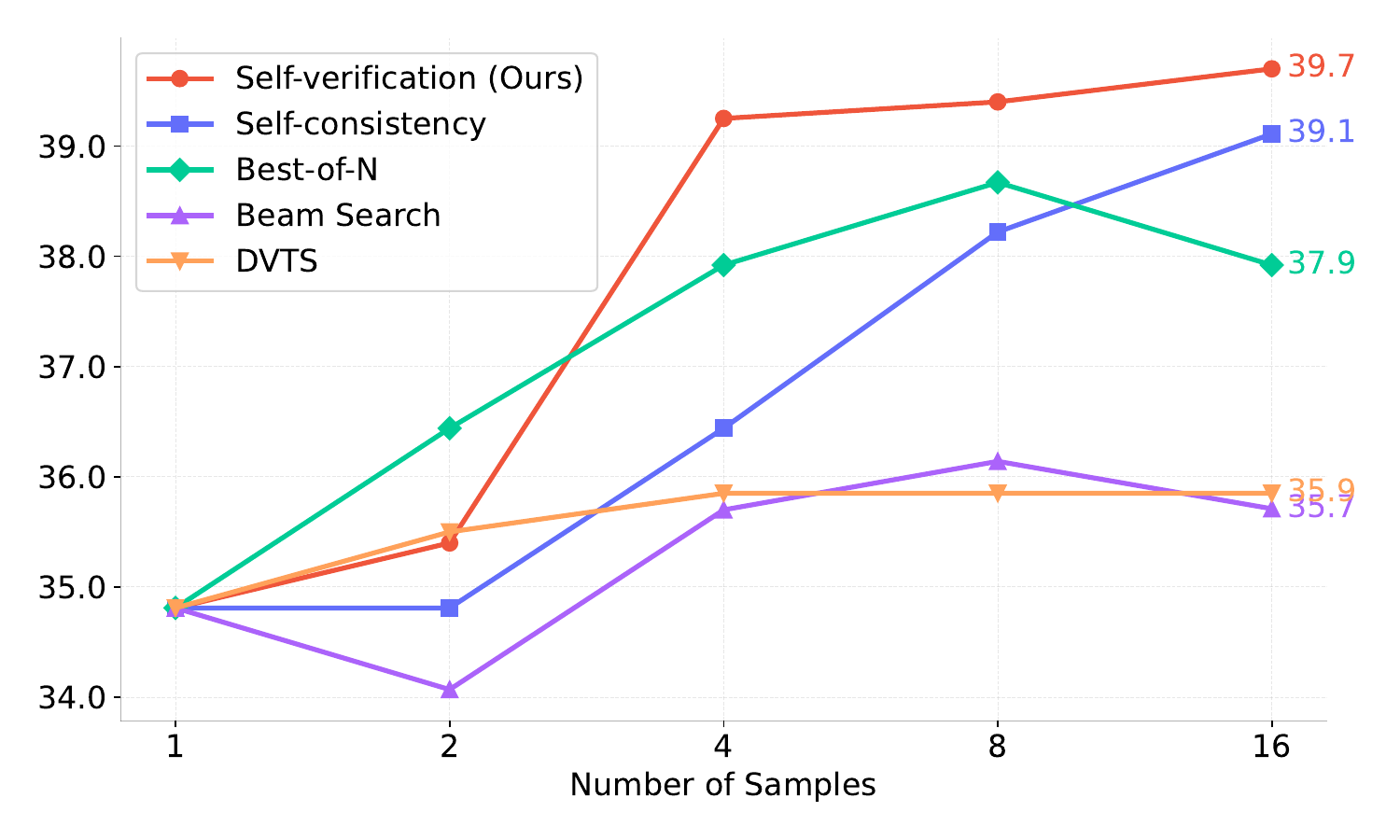}
        \caption{OlympiadBench Score (avg@1)}
        \label{fig:qwen-7b-olympiadbench}
    \end{subfigure}
    \caption{Test-time scaling performance of Self-Verification-Qwen-7B on math reasoning benchmarks including AIME24, AIME25, AMC23, and OlympiadBench.}
    \label{fig:qwen-7b-scaling}
\end{figure}

\begin{figure}[t]
    \centering
    \begin{subfigure}[t]{0.48\textwidth}
        \centering
        \includegraphics[width=\linewidth]{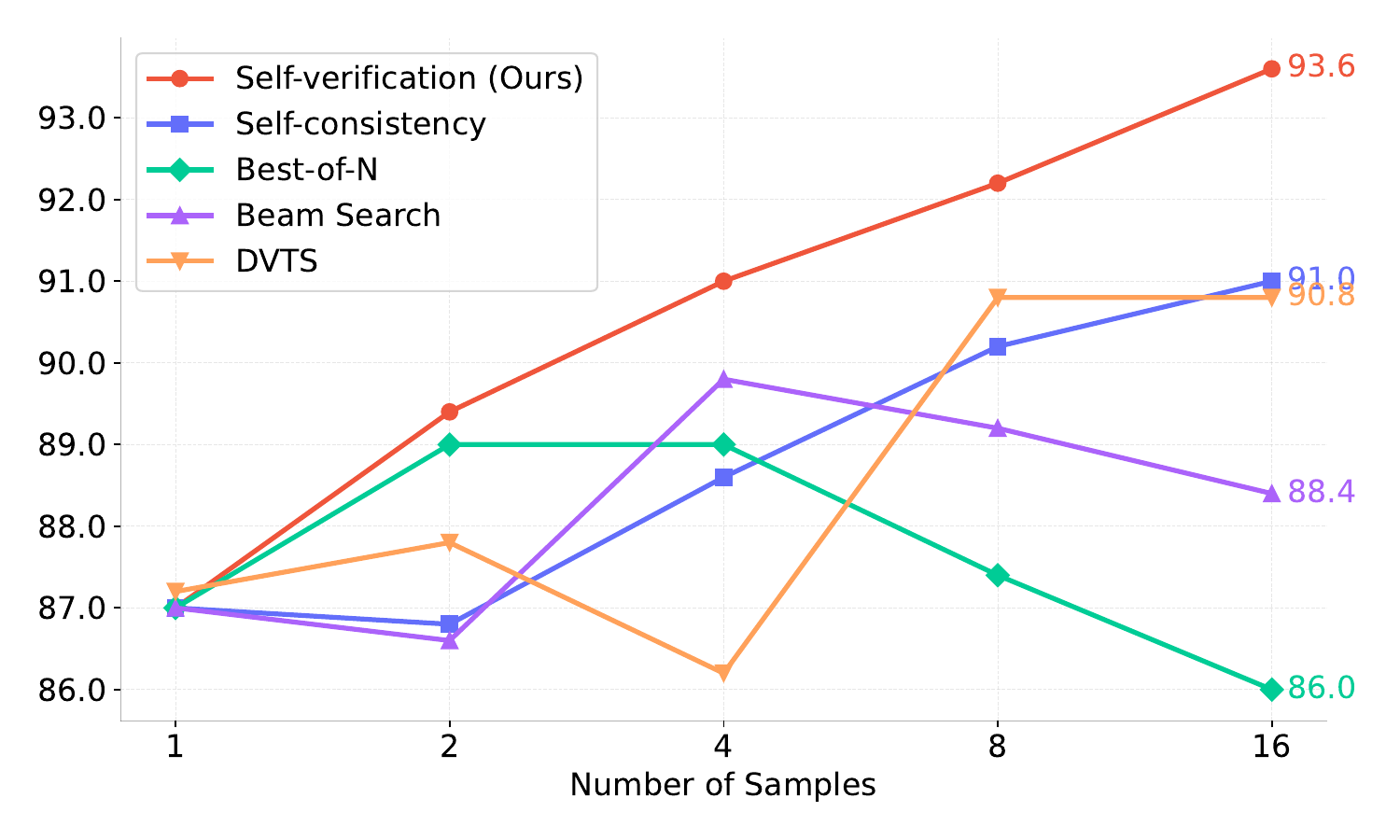}
        \caption{MATH500 Score (avg@1)}
        \label{fig:r1-1.5b-math500}
    \end{subfigure}
    \hfill
    \begin{subfigure}[t]{0.48\textwidth}
        \centering
        \includegraphics[width=\linewidth]{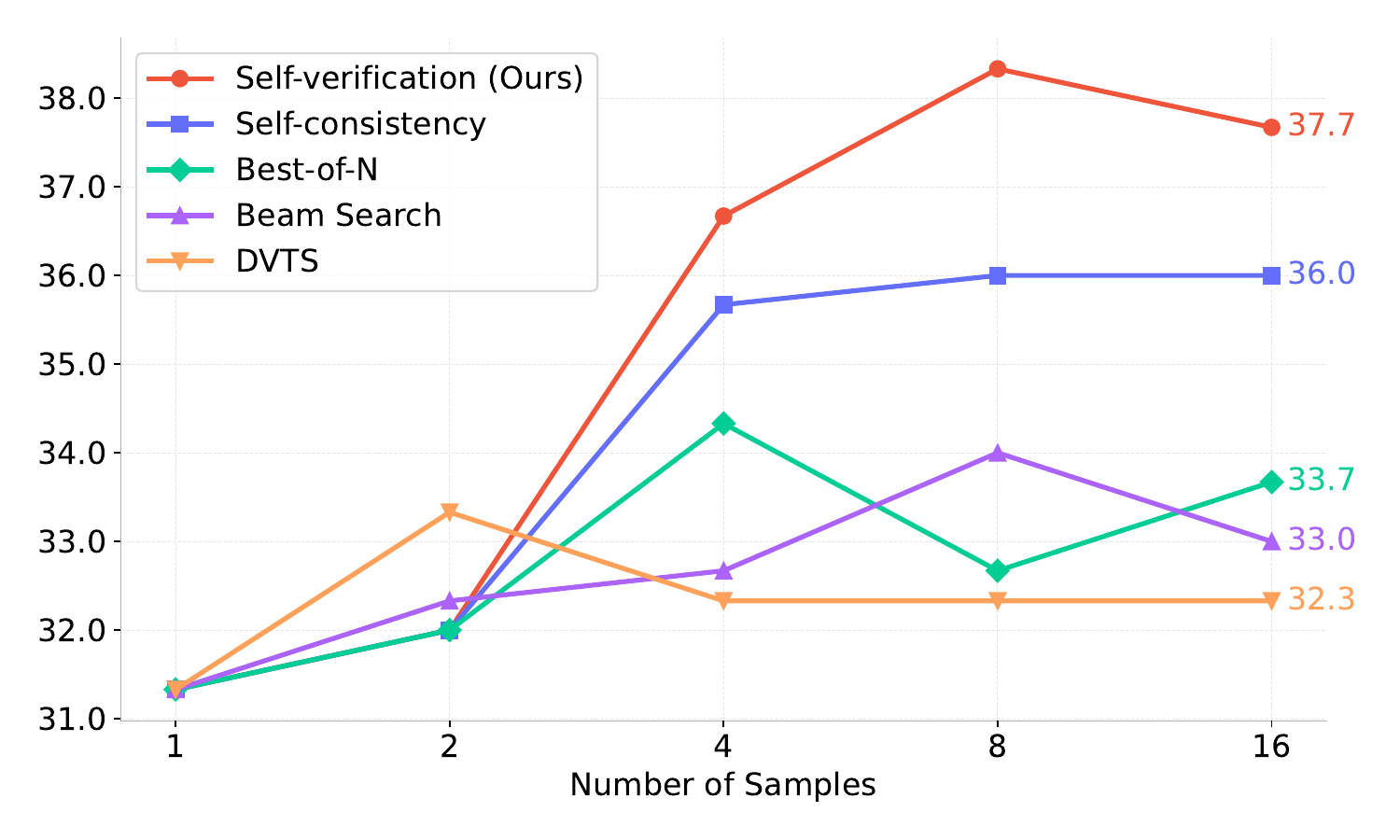}
        \caption{AIME25 Score (avg@10)}
        \label{fig:r1-1.5b-aime25}
    \end{subfigure}
    
    \begin{subfigure}[t]{0.48\textwidth}
        \centering
        \includegraphics[width=\linewidth]{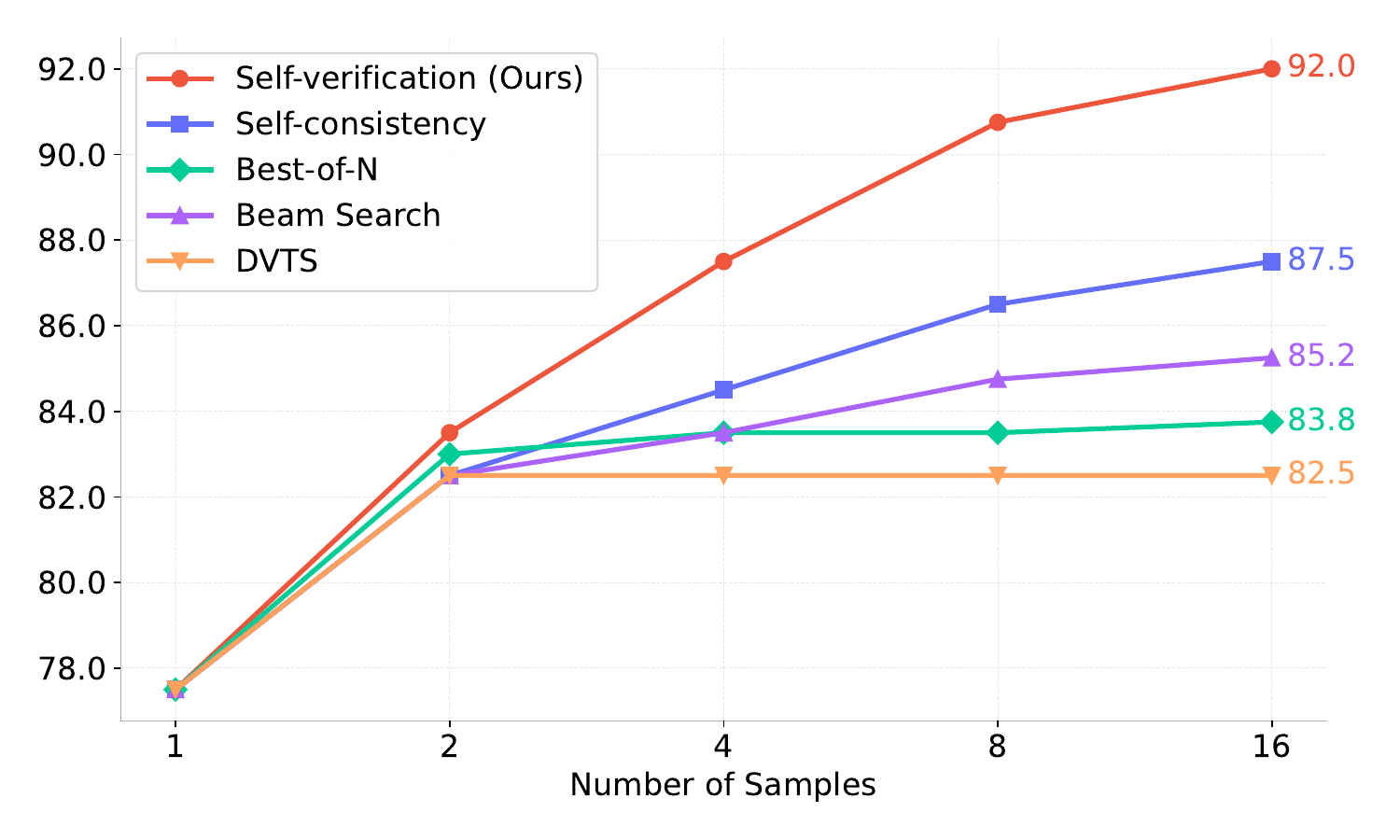}
        \caption{AMC23 Score (avg@10)}
        \label{fig:r1-1.5b-amc23}
    \end{subfigure}
    \hfill
    \begin{subfigure}[t]{0.48\textwidth}
        \centering
        \includegraphics[width=\linewidth]{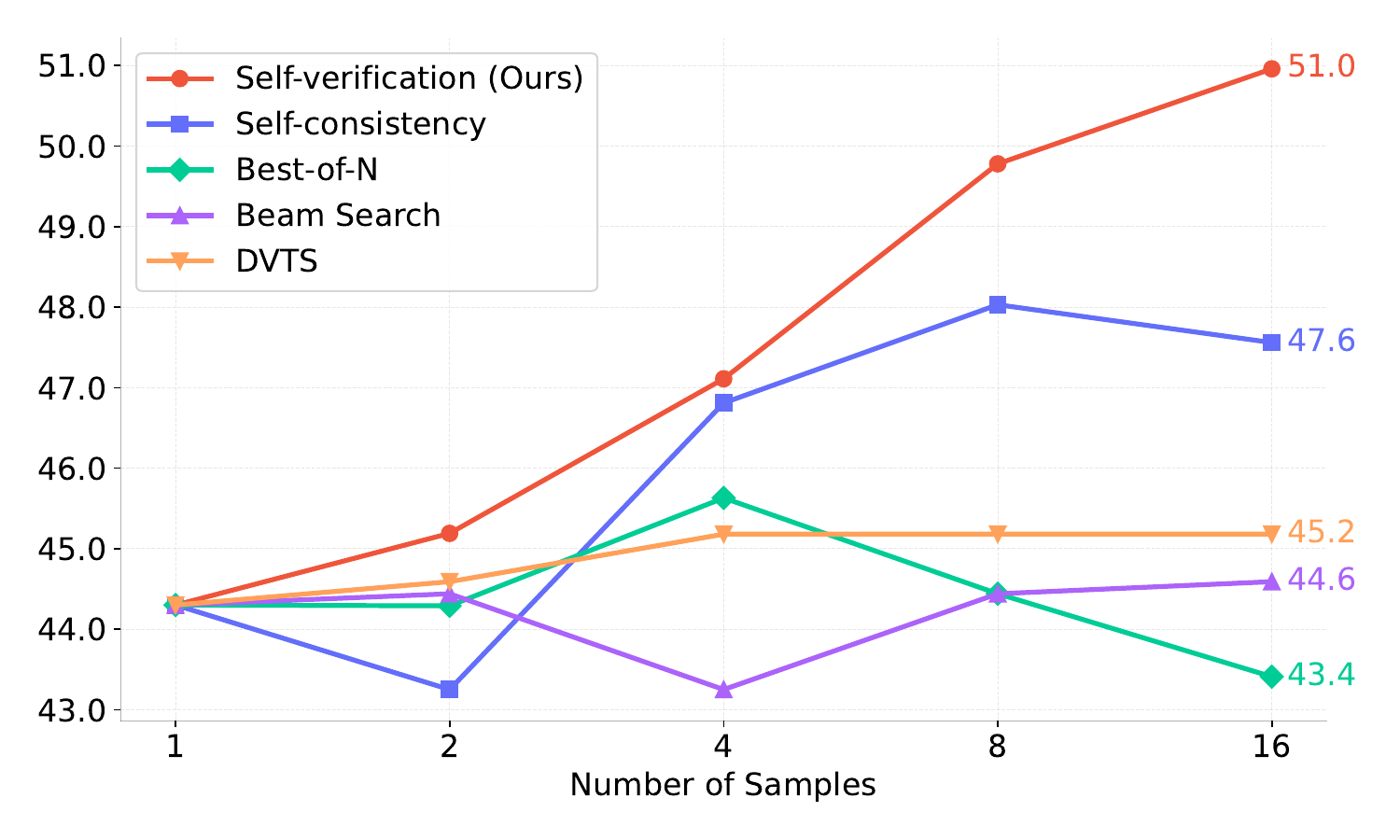}
        \caption{OlympiadBench Score (avg@1)}
        \label{fig:r1-1.5b-olympiadbench}
    \end{subfigure}
    \caption{Test-time scaling performance of Self-Verification-R1-1.5B on math reasoning benchmarks including MATH500, AIME25, AMC23, and OlympiadBench.}
    \label{fig:r1-1.5b-scaling}
\end{figure}

\section{Technical Details on Test-time Scaling}
\label{app:test-time-scaling}

In this section, we provide the details of our test-time scaling process. We implement an efficient inference-time generation framework on top of \citet{hf-test-time-scaling} but substitute the inference engine to SGLang \citep{sglang}, which provides a highly controllable frontend for guided generation. We use the same generation settings for all methods, whose hyperparameters are listed in \cref{tab:test-time-scaling-config}. We choose the RLHFlow \citep{rlhflow-prm} PRM for all baselines as their reward models. According to the practice of \citet{hf-test-time-scaling}, we use the last step score of the PRM as the reward for the current step or response. On the hyperparameter $\alpha$, which balances the answer selection between model verification score and the majority answer, we choose a relatively large value for Self-Verification-Qwen-7B model and a smaller value for Self-Verification-R1-1.5B. This is decided according to their verification accuracy during the training time, since we find the 7B model often has a better verification performance. In contrast, using smaller $\alpha$ value for the 1.5B model helps alleviate its verification error. For inference-time rollouts, we maintain the same temperature of $0.6$ as that in the training time. 

\paragraph{Computational Resources} We typically use 8 NVIDIA GPUs with 80GB memory each for all test-time scaling experiments. We first deploy the SGLang server on all GPUs with data parallel strategy. For baselines requiring simultaneous reward model deployment, we lower the GPU memory utilization accordingly to enable additional memory for the reward model. The inference of reward models are directly through the Hugging Face Transformers library since it is non-trivial to deploy the PRM forward pass on our inference engine. In contrast, the verification process of our self-verification model can be fully realized by the LLM generation process. As a result, we only need to serve one model simply through the deployed SGLang server, where we show the efficiency of our framework at test time in \cref{fig:time-cost}.

\section{Additional Results on Test-time Scaling}
\label{app:test-time-scaling-additional}

In \cref{fig:qwen-7b-scaling} and \cref{fig:r1-1.5b-scaling}, we show the test-time scaling performance of our Self-Verification-Qwen-7B and Self-Verification-R1-1.5B models on different math reasoning benchmarks, which corresponds to the results shown in the main paper. We find that our self-verification test-time method can achieve better performance than our compared baselines in most tasks. 

\section{Analysis on Distribution Discrepancy}
\label{app:distribution-discrepancy}

To substantiate our claim that a distribution discrepancy exists between post-trained generators and general reward models (RMs), we evaluate the calibration of an external RM on outputs from different generators. A well-aligned RM should assign higher scores to correct answers regardless of the generator model. We measure the Pearson correlation coefficient between the RM scores and ground-truth correctness on the MATH500 dataset. We use the \texttt{RLHFlow/Llama3.1-8B-PRM-Deepseek-Data} model as the external RM. As shown in \cref{tab:correlation-coefficient}, the correlation drops significantly for responses generated by our post-trained model compared to standard instruction-tuned models, indicating that the external RM is misaligned with our model's output distribution. This misalignment limits the effectiveness of test-time scaling methods that rely on such RMs.

\begin{table}[h]
    \centering
    \caption{Pearson correlation coefficient between external RM scores and ground-truth correctness on MATH500.}
    \label{tab:correlation-coefficient}
    \begin{tabular}{lc}
        \toprule
        \textbf{Generator Model} & \textbf{Pearson Correlation Coefficient} \\
        \midrule
        Llama-3.1-8B-Instruct & 44.9 \\
        Qwen2.5-Math-7B-Instruct & 42.3 \\
        Self-Verification-Qwen-7B & 37.5 \\
        \bottomrule
    \end{tabular}
\end{table}

\begin{figure}[t]
    \centering
    \begin{minipage}[t]{0.49\textwidth}
        \centering
        \includegraphics[width=\linewidth]{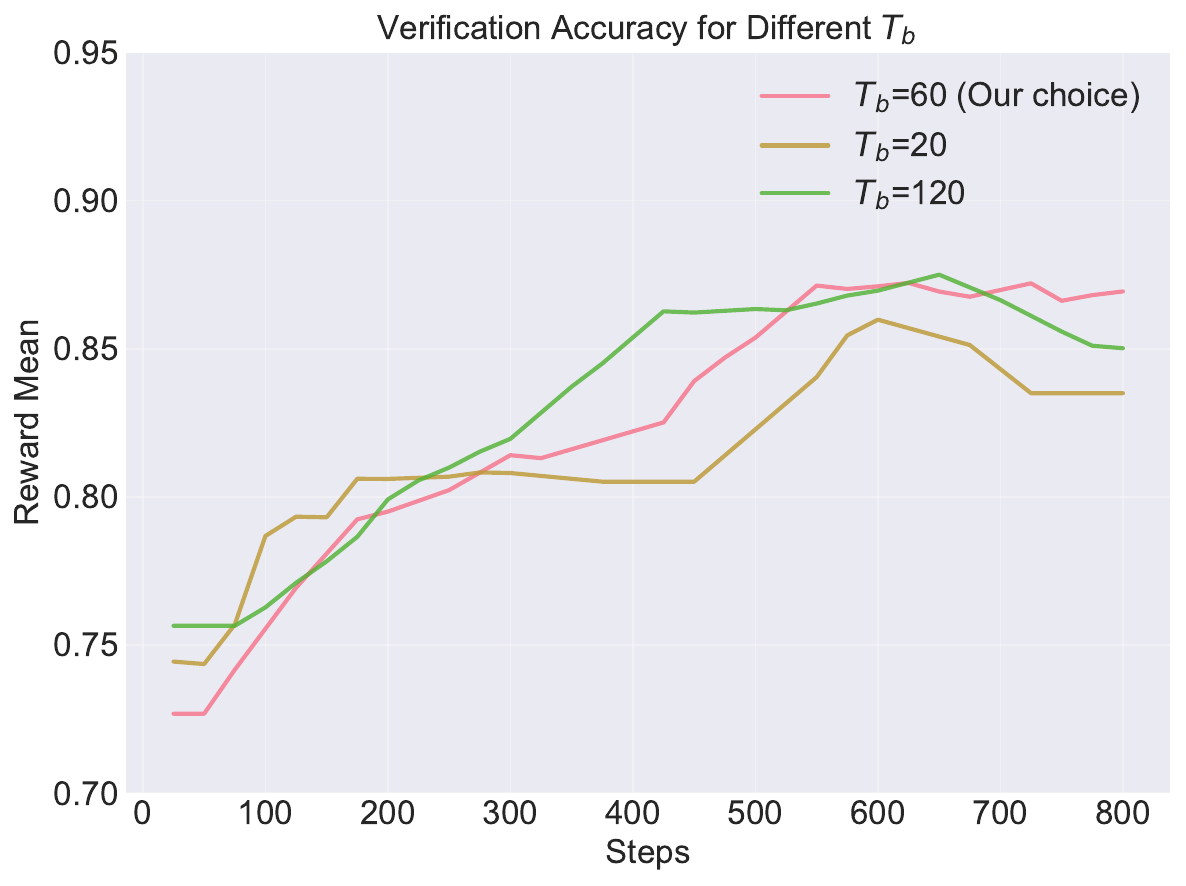}
        \caption{Verification accuracy on different choices of the policy-aligned buffer update frequency $T_b$ from the Qwen2.5-Math-7B base model. The values are smoothed with a window size of 5.}
        \label{fig:buffer-step-comparison}
    \end{minipage}
    \hfill
    \begin{minipage}[t]{0.49\textwidth}
        \centering
        \includegraphics[width=\linewidth]{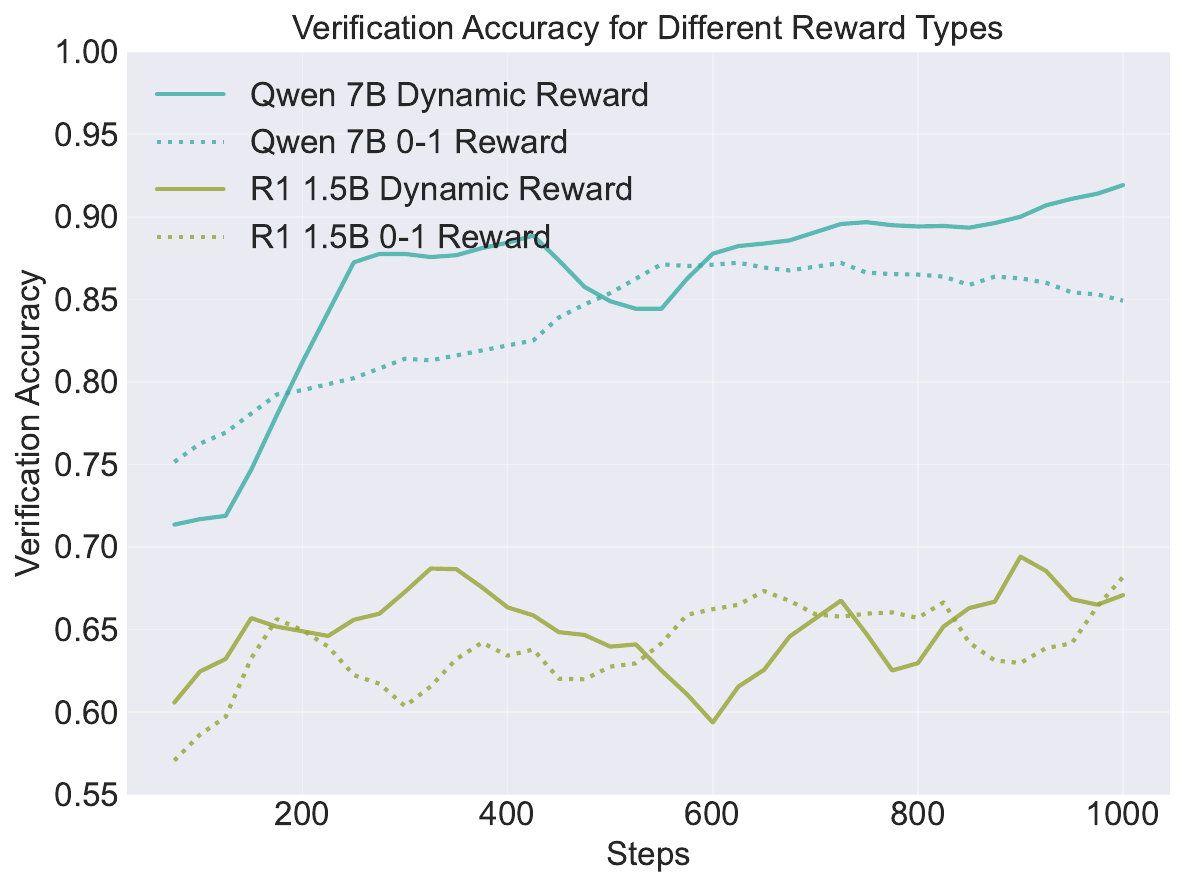}
        \caption{Verification accuracy difference between the dynamic reward design and a simple 0-1 reward for RL with Qwen 7B and R1 1.5B models. The values are smoothed with a window size of 5.}
        \label{fig:reward-type-comparison}
    \end{minipage}
\end{figure}

\section{Ablation Studies}
\label{app:ablation-studies}

We investigate how our framework's core design choices influence verification quality. First, we examine the policy-aligned buffer, which keeps the verifier on-policy by refreshing stored trajectories every $T_b$ steps. As shown in \cref{fig:buffer-step-comparison}, the verification accuracy curves remain tightly clustered when sweeping $T_b$ around our default value. While larger buffers amortize variance, they provide little extra signal during the online training process and thus show lower accuracy at later training stages. In contrast, $T_b=20$ shows less stability during the training process. We therefore retain $T_b = 60$ as a robust trade-off. 

Next, we analyze the dynamic verification reward, which reweights samples by their inferred difficulty. \Cref{fig:reward-type-comparison} plots the difference in verification accuracy between using the dynamic reward and a simple binary reward. The figure shows that for the 7B model, the dynamic reward yields a significant and sustained positive accuracy difference, indicating a clear performance improvement. For the 1.5B model, the gain is less pronounced throughout training. This suggests that the smaller model's limited capacity may prevent it from fully leveraging the nuanced signal from the difficulty-aware reward. The dynamic reward is efficient in our implementation since the difficulty statistics are derived from GRPO rollouts already generated during training.

\section{Comparison with Entropy-Minimization Baseline}
\label{app:em-baseline}

Recent work has shown that entropy minimization (EM) can be a strong baseline for post-training LLMs. We compare our Self-Verification-Qwen-7B model with the one-shot entropy minimization model (EM 1-shot) from \citet{gao2025oneshot}, which is also post-trained from the Qwen2.5-Math-7B base model. As shown in \cref{tab:em-comparison}, our model outperforms the EM baseline across all benchmarks, demonstrating the effectiveness of our joint problem-solving and verification framework.

\begin{table}[h]
    \centering
    \caption{Comparison with the entropy-minimization (EM) baseline on greedy-decoding scores.}
    \label{tab:em-comparison}
    \begin{tabular}{lccccc}
    \toprule
    Model & MATH500 & \makecell{AIME24\\(avg@10)} & \makecell{AIME25\\(avg@10)} & \makecell{AMC23\\(avg@10)} & \makecell{Olympiad\\Bench} \\
    \midrule
    Self-Verification-Qwen-7B & 83.6 & 20.00 & 16.67 & 63.75 & 34.81 \\
    EM 1-shot & 68.8 & 19.33 & 8.67 & 54.50 & 27.26 \\
    \bottomrule
    \end{tabular}
\end{table}

\section{Prompting on Generative Verification}
\label{app:verification-prompting}

Our generative verification process follows the previous work \citep{generative-verifier}. Additionally, we adopt a multi-turn chat template to generate the verification output. A template is shown in the box below. We first append the problem and solution for the given data sample, and then add the verification prompt in a fixed format.

\begin{lstlisting}[caption=Generative Verification Prompting Template,label={lst:verification-template}, basicstyle=\ttfamily]
User: [Math problem]

Assistant: [Solution]

User: Please verify the solution step by step. At the end of the solution verification, when you give your final grade, write it in the form "Is the answer correct (Yes/No)? X", where X is either Yes or No.

Assistant: 
\end{lstlisting}

With this template, the LLM is prone to answer the verification question with `Yes' or `No' at the end. When performing generation with inference engine, we can simply set the stopping string at this template and thus we can acquire the probability of the `Yes' token according to \cref{eq:verifier-score}.

\section{Examples of Problem-Solving and Verification Outputs}
\label{app:output-examples}

In this section, we provide some generation examples on problem-solving and verification tasks from our Self-Verification-Qwen-7B (Listing~\ref{lst:problem-solving-example-qwen-7b} and Listing~\ref{lst:verification-example-qwen-7b}) and Self-Verification-R1-1.5B models (Listing~\ref{lst:problem-solving-example-r1-1.5b} and Listing~\ref{lst:verification-example-r1-1.5b}). 

\begin{lstlisting}[caption=Self-Verification-Qwen-7B Problem-Solving Example,label={lst:problem-solving-example-qwen-7b}, basicstyle=\ttfamily\scriptsize]

    User: Suppose that I have $6$ different books, $2$ of which are math books. In how many ways can I stack my $6$ books on a shelf if I do not want the math books to be next to each other?
    
    Assistant: To determine the number of ways to stack 6 different books on a shelf such that the 2 math books are not next to each other, we can use the principle of complementary counting. This involves calculating the total number of possible arrangements of the books and then subtracting the number of arrangements where the math books are next to each other.
    
    **Step 1: Calculate the total number of arrangements of the 6 books.**
    
    The total number of ways to arrange 6 different books is given by the factorial of 6:
    \[
    6! = 6 \times 5 \times 4 \times 3 \times 2 \times 1 = 720
    \]
    
    **Step 2: Calculate the number of arrangements where the 2 math books are next to each other.**
    
    To treat the 2 math books as a single unit, we can think of them as forming a "super book." This reduces the problem to arranging 5 "books" (the "super book" and the other 4 individual books). The number of ways to arrange these 5 "books" is:
    \[
    5! = 5 \times 4 \times 3 \times 2 \times 1 = 120
    \]
    Within the "super book," the 2 math books can be arranged in 2 different ways (either math book 1 followed by math book 2, or math book 2 followed by math book 1). Therefore, the total number of arrangements where the 2 math books are next to each other is:
    \[
    5! \times 2 = 120 \times 2 = 240
    \]
    
    **Step 3: Subtract the number of arrangements where the math books are next to each other from the total number of arrangements.**
    
    The number of arrangements where the 2 math books are not next to each other is:
    \[
    6! - 5! \times 2 = 720 - 240 = 480
    \]
    
    Thus, the number of ways to stack the 6 books on a shelf such that the math books are not next to each other is \(\boxed{480}\).
    
    \end{lstlisting}
    
    \begin{lstlisting}[caption=Self-Verification-Qwen-7B Verification Example,label={lst:verification-example-qwen-7b},basicstyle=\ttfamily\scriptsize]
    
    User: If $a$ is three times $b$, $b$ is twice $c$, and $c$ is four times $d$, what is the value of $\frac{a\cdot c}{b\cdot d}$?
    
    Assistant: To find the value of \(\frac{a \cdot c}{b \cdot d}\), we start by expressing \(a\), \(b\), and \(c\) in terms of \(d\).
    
    Given:
    1. \(a = 3b\)
    2. \(b = 2c\)
    3. \(c = 4d\)
    
    First, substitute \(c = 4d\) into \(b = 2c\):
    \[b = 2(4d) = 8d.\]
    
    Next, substitute \(b = 8d\) into \(a = 3b\):
    \[a = 3(8d) = 24d.\]
    
    Now we have \(a = 24d\), \(b = 8d\), and \(c = 4d\). We substitute these expressions into \(\frac{a \cdot c}{b \cdot d}\):
    \[
    \frac{a \cdot c}{b \cdot d} = \frac{(24d) \cdot (4d)}{(8d) \cdot d} = \frac{96d^2}{8d^2} = \frac{96}{8} = 12.
    \]
    
    Therefore, the value is \(\boxed{12}\).
    
    User: Please verify the solution step by step. At the end of the solution verification, when you give your final grade, write it in the form "Is the answer correct (Yes/No)? X", where X is either Yes or No.
    
    Assistant: To verify the solution, let's re-evaluate the expression step by step.
    
    Given:
    1. \(a = 3b\)
    2. \(b = 2c\)
    3. \(c = 4d\)
    
    First, express \(b\) in terms of \(d\):
    \[b = 2c = 2(4d) = 8d.\]
    
    Next, express \(a\) in terms of \(d\):
    \[a = 3b = 3(8d) = 24d.\]
    
    Now, substitute \(a = 24d\), \(b = 8d\), and \(c = 4d\) into the expression \(\frac{a \cdot c}{b \cdot d}\):
    \[
    \frac{a \cdot c}{b \cdot d} = \frac{(24d) \cdot (4d)}{(8d) \cdot d} = \frac{96d^2}{8d^2} = \frac{96}{8} = 12.
    \]
    
    The value is indeed \(\boxed{12}\).
    
    Is the answer correct (Yes/No)? Yes
    \end{lstlisting}
    
    \begin{lstlisting}[caption=Self-Verification-R1-1.5B Problem-solving Example,label={lst:problem-solving-example-r1-1.5b},basicstyle=\ttfamily\scriptsize]
    
    User: Suppose that I have $6$ different books, $2$ of which are math books. In how many ways can I stack my $6$ books on a shelf if I do not want the math books to be next to each other?
    
    Assistant: (We omit the thinking part since it is too long.)
    
    First, we calculate the total number of ways to arrange 6 different boosks without any restrictions. This is given by \(6!\):
    
    \[
    6! = 720
    \]
    
    Next, we need to subtract the number of arrangements where the two math books are next to each other. To find this, we treat the two math books as a single unit or "block." This reduces the problem to arranging 5 "books" (the "super book" and the other 4 individual books). The number of ways to arrange these 5 "books" is:
    \[
    5! = 5 \times 4 \times 3 \times 2 \times 1 = 120
    \]
    Within the "super book," the two math books can be arranged in 2 different ways (either math book 1 followed by math book 2, or math book 2 followed by math book 1). Therefore, the total number of arrangements where the 2 math books are next to each other is:
    \[
    5! \times 2 = 120 \times 2 = 240
    \]
    
    Subtracting the number of bad arrangements from the total number of arrangements gives us the number of valid arrangements where the math books are not next to each other:
    
    \[
    720 - 240 = 480
    \]
    
    Thus, the number of ways to stack the 6 books on a shelf such that the math books are not next to each other is \(\boxed{480}\).
    
    
    \end{lstlisting}
    
    \begin{lstlisting}[caption=Self-Verification-R1-1.5B Verification Example,label={lst:verification-example-r1-1.5b},basicstyle=\ttfamily\scriptsize]
    
    User: Given that events A and B are independent, and both are mutually exclusive with event C. It is known that $P(A) = 0.2$, $P(B) = 0.6$, and $P(C) = 0.14$. Find the probability that at least one of A, B, or C occurs, denoted as $P(A+B+C)$.
    
    Assistant: Given that events A and B are independent, and both are mutually exclusive with event C, we need to find the probability that at least one of A, B, or C occurs, denoted as \( P(A \cup B \cup C) \).
    
    We know the following probabilities:
    - \( P(A) = 0.2 \)
    - \( P(B) = 0.6 \)
    - \( P(C) = 0.14 \)
    
    Since A and B are independent, the probability of their intersection is:
    \[ P(A \cap B) = P(A) \times P(B) = 0.2 \times 0.6 = 0.12 \]
    
    Since A and B are mutually exclusive with C, the probabilities of A and C overlapping, and B and C overlapping, are both zero:
    \[ P(A \cap C) = 0 \]
    \[ P(B \cap C) = 0 \]
    
    Using the inclusion-exclusion principle for three events, we have:
    \[ P(A \cup B \cup C) = P(A) + P(B) + P(C) - P(A \cap B) - P(A \cap C) - P(B \cap C) + P(A \cap B \cap C) \]
    
    Substituting the known values:
    \[ P(A \cup B \cup C) = 0.2 + 0.6 + 0.14 - 0.12 - 0 - 0 + 0 \]
    \[ P(A \cup B \cup C) = 0.2 + 0.6 + 0.14 - 0.12 \]
    \[ P(A \cup B \cup C) = 0.94 - 0.12 \]
    \[ P(A \cup B \cup C) = 0.82 \]
    
    Thus, the probability that at least one of A, B, or C occurs is \(\boxed{0.82}\).
    
    User: Please verify the solution step by step. At the end of the solution verification, when you give your final grade, write it in the form "Is the answer correct (Yes/No)? X", where X is either Yes or No.
    
    Assistant: 
    
    <think>
    
    Okay, so I need to verify if my answer of 0.82 is correct. Let me think through this step by step.
    
    First, I recall that events A and B are independent. That means the probability of both A and B happening together is the product of their individual probabilities. So, P(A ∩ B) should be P(A) * P(B) = 0.2 * 0.6 = 0.12. That seems right.
    
    Next, events A and B are mutually exclusive with C. That means A and C can't happen at the same time, and B and C can't happen at the same time. So, P(A ∩ C) = 0 and P(B ∩ C) = 0. Okay, that makes sense.
    
    Now, I used the inclusion-exclusion principle for three events to find P(A ∪ B ∪ C). The formula is:
    
    P(A ∪ B ∪ C) = P(A) + P(B) + P(C) - P(A ∩ B) - P(A ∩ C) - P(B ∩ C) + P(A ∩ B ∩ C)
    
    Plugging in the numbers:
    
    P(A ∪ B ∪ C) = 0.2 + 0.6 + 0.14 - 0.12 - 0 - 0 + 0
    
    Wait a minute, why are P(A ∩ C) and P(B ∩ C) zero? Because the events are mutually exclusive with C, so yes, they shouldn't overlap. But is there a chance that all three events A, B, and C might overlap? That is, is P(A ∩ B ∩ C) not zero?
    
    Hmm, the problem says A and B are mutually exclusive with C, which I think means that A cannot occur with C, and B cannot occur with C. But A and B can happen separately from C.
    
    So, perhaps the total probability is P(A) + P(B) + P(C) - P(A ∩ B) - P(A ∩ C) - P(B ∩ C) + P(A ∩ B ∩ C). But since P(A ∩ C) is zero, and P(B ∩ C) is zero, and P(A ∩ B) is 0.12, and P(A ∩ B ∩ C) is also zero,
    
    Wait, maybe I'm confusing things.
    
    If A and B are independent, then P(A ∩ B) = P(A) * P(B) = 0.12. Since A and B are mutually exclusive with C, that means P(A ∩ C) = 0 and P(B ∩ C) = 0. But what about P(A ∩ B ∩ C)? Since if A ∩ C is zero, and B ∩ C is zero, then A ∩ B ∩ C would also be zero because it's a subset of both A ∩ C and B ∩ C.
    
    So, P(A ∩ B ∩ C) = 0.
    
    Therefore, the formula simplifies to:
    
    P(A ∪ B ∪ C) = P(A) + P(B) + P(C) - P(A ∩ B) - P(A ∩ C) - P(B ∩ C) + P(A ∩ B ∩ C)
    
    Plugging in the numbers:
    
    = 0.2 + 0.6 + 0.14 - 0.12 - 0 - 0 + 0
    
    = 0.2 + 0.6 + 0.14 - 0.12
    
    = 0.94 - 0.12
    
    = 0.82
    
    So, the calculation seems correct. Therefore, the answer of 0.82 is accurate.
    
    </think>
    
    Is the answer correct (Yes/No)? Yes
    
    
    
\end{lstlisting}

\end{document}